\documentclass[manuscript,screen]{acmart}
\usepackage{xcolor}
\usepackage{soul}
\usepackage{graphicx}
\usepackage[ruled,linesnumbered]{algorithm2e}
\usepackage{amsmath}
\usepackage{amsmath}
\usepackage{color,soul}
\usepackage{caption}
\usepackage{subcaption}
\usepackage[normalem]{ulem}
\usepackage{multirow}
\usepackage{multicol}
\usepackage{amsfonts}
\usepackage{url}

\DeclareMathOperator*{\argmin}{arg\,min}

\definecolor{purple}{RGB}{210, 0, 210} %makes good comment color
\definecolor{green}{RGB}{0, 153, 0} %makes good comment color
\AtBeginDocument{%
  \providecommand\BibTeX{{%
    \normalfont B\kern-0.5em{\scshape i\kern-0.25em b}\kern-0.8em\TeX}}}

\setcopyright{acmcopyright}
\copyrightyear{2022}
\acmYear{2022}
\acmDOI{XXXXXXX.XXXXXXX}

\begin{document}

%%
%% The "title" command has an optional parameter,
%% allowing the author to define a "short title" to be used in page headers.
\title{Occlusion-Robust Multi-Sensory Posture Estimation in Physical Human-Robot Interaction}

%%
%% The "author" command and its associated commands are used to define
%% the authors and their affiliations.
%% Of note is the shared affiliation of the first two authors, and the
%% "authornote" and "authornotemark" commands
%% used to denote shared contribution to the research.
% \author{Amir Yazdani}
% % \authornote{Both authors contributed equally to this research.}
% \email{amir.yazdani@utah.edu}
% \orcid{0000-0002-1920-8456}
\author{Amir Yazdani}
\orcid{0000-0002-1920-8456}
\authornotemark[1]
\email{amir.yazdani@utah.edu}
\affiliation{%
  \institution{University of Utah Robotics Center}
%   \streetaddress{P.O. Box 1212}
  \city{Salt Lake City}
  \state{Utah}
  \country{USA}
%   \postcode{43017-6221}
}

\author{Roya Sabbagh Novin}
\orcid{0000-0003-4265-0661}
\authornotemark[1]
\email{roya.sabbaghnovin@utah.edu}
\affiliation{%
  \institution{University of Utah Robotics Center}
%   \streetaddress{P.O. Box 1212}
  \city{Salt Lake City}
  \state{Utah}
  \country{USA}
%   \postcode{43017-6221}
}

\author{Andrew Merryweather}
\orcid{0000-0002-9048-9473}
\email{a.merryweather@utah.edu}
\authornotemark[1]
\affiliation{%
  \institution{University of Utah Robotics Center}
%   \streetaddress{P.O. Box 1212}
  \city{Salt Lake City}
  \state{Utah}
  \country{USA}
%   \postcode{43017-6221}
}
\authornotemark[2]
\affiliation{%
  \institution{Department of Mechanical Engineering}
%   \streetaddress{P.O. Box 1212}
  \city{Salt Lake City}
  \state{Utah}
  \country{USA}
%   \postcode{43017-6221}
}

\author{Tucker Hermans}
\orcid{0000-0003-2496-2768}
\email{tucker.hermans@utah.edu}
\authornotemark[1]
\affiliation{%
  \institution{University of Utah Robotics Center}
%   \streetaddress{P.O. Box 1212}
  \city{Salt Lake City}
  \state{Utah}
  \country{USA}
%   \postcode{43017-6221}
}
\authornotemark[2]
\affiliation{%
  \institution{NVIDIA}
%   \streetaddress{P.O. Box 1212}
  \city{Seattle}
  \state{Washington}
  \country{USA}
%   \postcode{43017-6221}
}

%%
%% By default, the full list of authors will be used in the page
%% headers. Often, this list is too long, and will overlap
%% other information printed in the page headers. This command allows
%% the author to define a more concise list
%% of authors' names for this purpose.
\renewcommand{\shortauthors}{Yazdani, et al.}

%%
%% The abstract is a short summary of the work to be presented in the
%% article.
\begin{abstract}
3D posture estimation is important in analyzing and improving ergonomics in physical human-robot interaction and reducing the risk of musculoskeletal disorders. Vision-based posture estimation approaches are prone to sensor and model errors, as well as occlusion, while posture estimation solely from the interacting robot's trajectory suffers from ambiguous solutions. To benefit from the advantages of both approaches and improve upon their drawbacks, we introduce a low-cost, non-intrusive, and occlusion-robust multi-sensory 3D postural estimation algorithm in physical human-robot interaction. We use 2D postures from OpenPose over a single  camera, and the trajectory of the interacting robot while the human performs a task. 
We model the problem as a partially-observable dynamical system and we infer the 3D posture via a particle filter. We present our work in teleoperation, but it can be generalized to other applications of physical human-robot interaction. We show that our multi-sensory system resolves human kinematic redundancy better than posture estimation solely using OpenPose or posture estimation solely using the robot's trajectory. This will increase the accuracy of estimated postures compared to the gold-standard motion capture postures. Moreover, our approach also performs better than other single sensory methods when postural assessment using RULA assessment tool.
\end{abstract}

%%
%% The code below is generated by the tool at http://dl.acm.org/ccs.cfm.
%% Please copy and paste the code instead of the example below.
%%


%%
%% Keywords. The author(s) should pick words that accurately describe
%% the work being presented. Separate the keywords with commas.
%   \begin{teaserfigure}
%     \includegraphics[width=\textwidth]{figures/ergo_intelligence_framework.png}
%     \caption{figure caption}
%     \Description{figure description}
%   \end{teaserfigure}

%%
%% This command processes the author and affiliation and title
%% information and builds the first part of the formatted document.
\maketitle

\section{Introduction}
% \OL{- Why 3d posture estimation is important?}\\
3D human posture estimation is a key component in human-related sciences such as biomechanics, ergonomics, sport medicine, entertainments, game development, and human-robot interaction. It requires a sufficiently-accurate and robust motion tracker which can be easily used to evaluate human movements~\cite{corazza2006markerless}. Traditionally, marker-based motion capture systems~(MoCap) were the main approach in biomechanics studies due to their high frequency of operation and accuracy~\cite{needham2021accuracy}. However, due to their personal, environmental, and technical constraints~\cite{topley2020comparison,seethapathi2019movement,colyer2018review,kessler2019direct, tsushima2003test, metcalf2020quantifying}, new markerless methods are being developed that benefit from deep learning techniques to extract human posture from images and videos~\cite{cao2017realtime,fang2017rmpe,mathis2018deeplabcut}.
Among them, OpenPose~\cite{cao2019openpose} is a popular markerless method for human motion analysis due to its ease of use and integration.
% \OL{- what are the challenges in 3D posture estimation?}\\

Markerless techniques are more adaptable, minimally intrusive, and less expensive than marker-based systems. However, they require time-consuming calibration to deal with errors and uncertainties from the sensors~\cite{xiao2018wearable,chen2000camera}. Vision-based markerless methods can also be perturbed by the lighting, background color and even the user's clothing~\cite{erol2007vision} which will cause inaccurate segment lengths and joint angles. More importantly, occlusion is a significant issue which leads to loosing data and inability to track human motion and collect accurate and continuous posture data~\cite{steinebach2020accuracy,busch2017postural}. In posture estimation, occlusion might happen by an object in the environment masking a part of the camera view, or a human body part masking other segments of the body~(self-occlusion)~\cite{slembrouck2020multiview}. Although using multiple cameras with optimal placements~\cite{beriault2008multi} can reduce occlusion, it requires a more sophisticated and expensive system, which needs to be tailored for specific working environments and applications. 

% \OL{- Motivation from MSDs in Teleoperation}\\
One important application of 3D human posture estimation is human motion assessment in workplaces to reduce the risk of work-related musculoskeletal disorders~(WMSDs)~\cite{punnett2004work,erdinc2008ergonomics}. WMSDs are the second-largest source of disabilities worldwide~\cite{vos2015global}, and awkward postures are the main contributor to the work-related injuries~\cite {schneider2001musculoskeletal,da2010risk,yu2012work}. Teleoperation, as an example of physical human-robot interaction~(pHRI), is a well suited alternative for high-risk tasks~(e.g. construction and handling hazardous materials), since the remote workstation for the human can be designed ergonomically~\cite{dempsey2018emerging}. However, WMSDs are still common among human operators, even when they perform teleoperation without force feedback~\cite{peternel2020human, yu2014ergonomic}. 

\begin{figure}[t!]
    \includegraphics[width=\textwidth]{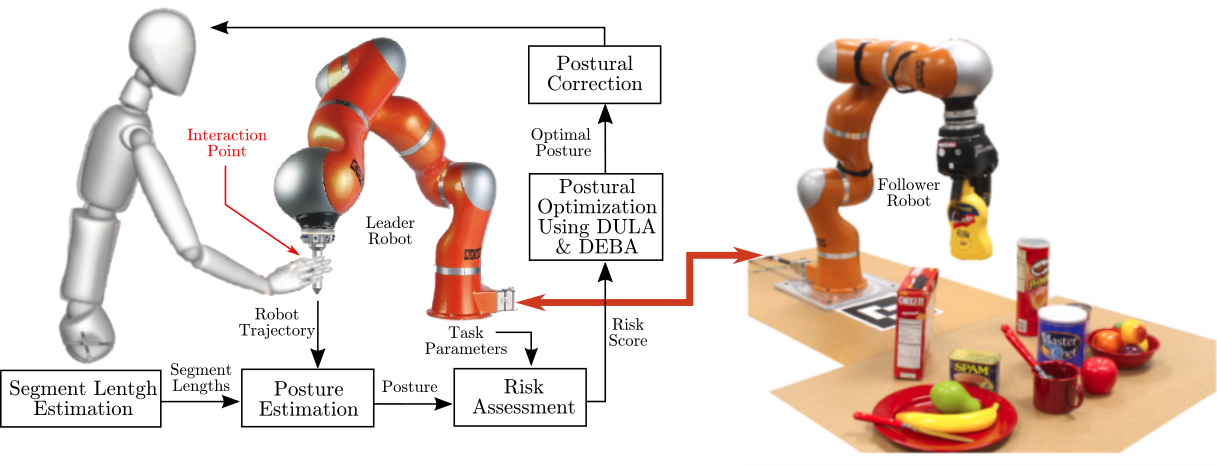}
    \caption{The framework for ergonomically intelligent teleoperation system. The same framework is true for other pHRI applications just by removing the follower robot.}
    \label{fig:framework}
  \end{figure}

In teleoperation, a human operator remotely controls the follower robot using the leader robot (see Fig.~\ref{fig:framework}). Using the leader robot in such proximity to the human increases occlusion and reduces the accuracy of posture estimation via markerless methods as mentioned in~\cite{busch2017postural}. However, it also provides the ability to use the leader robot stylus as an additional sensor as we initially proposed in~\cite{yazdani2020leader}.
In this paper, we focus on this teleoperation application of pHRI, however, our proposed approach and implementations can be extended with minor modification to other application with direct human and robot interactions.

% \OL{- Big picture of ergonomically intelligent teleoperation, focusing on posture estimation and assessment and combining with openpose}\\
To improve ergonomics and lower the risk of WMSDs in teleoperation, we introduced a {\itshape ergonomically intelligent teleoperation System} in~\cite{yazdani2021posture,yazdani2021ergonomically}, in which the teleoperation system intelligently estimates the human posture during interaction with the leader robot~(more details in~\cite{yazdani2020leader}), assesses the posture using ergonomics risk assessment tools, and if the risk of WMSDs is high, provides an ergonomically optimal postural correction for the human to complete the task~(more details in~\cite{yazdani2022dula}). We also introduced different types of autonomous postural correction in teleoperation using the leader robot. Figure~\ref{fig:framework} illustrates the details of the framework.

In this paper, we focus on improving the posture estimation performance to assess the ergonomics and risk of WMSDs in teleoperation tasks. Previously, in~\cite{yazdani2020leader}, we showed that the leader robot is an adequate sensor for posture estimation in teleoperation and we can use high resolution sensory information from the robot encoders for continuous, low-cost and occlusion-free monitoring of the 3D posture and risk of WMSDs in teleoperation with an adequate accuracy. However, to deal with the under-determined and ambiguous solutions, we had to limit the torso range of motion based on prior information from the type of tasks, and initialize the estimation approach around the neutral posture. Here, we improve upon our previous approach by combining it with a markerless posture estimation approach to provide occlusion-robust posture estimation with improved redundancy resolution.

\begin{figure}[t!]
    \includegraphics[width=\textwidth]{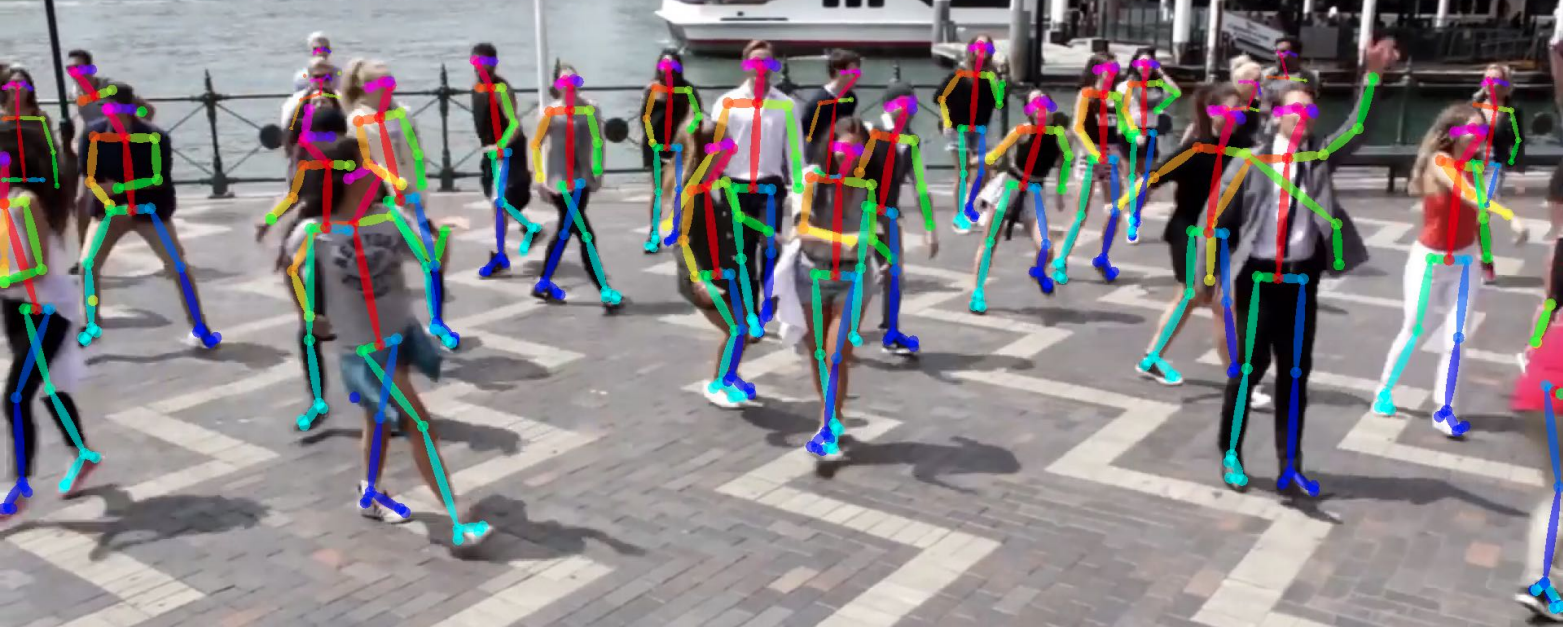}
    \caption{An example of estimated postures from OpenPose~\cite{cao2019openpose}.}
    \label{fig:openpose_example}
  \end{figure}

We use the popular OpenPose~\cite{cao2019openpose} method with the input of single-view video frames from a RGB camera as our additional sensory information. OpenPose is a markerless posture estimation approach which estimates the 2D postures from a single camera. They use a non-parametric representation, which they refer to it as part affinity fields, to learn to associate body parts with individuals in the image. Figure~\ref{fig:openpose_example} shows an example of estimated 2D postures from OpenPose. Although OpenPose estimates the postures in an online form, it suffers from sensor uncertainty, model accuracy, and occlusion~\cite{d2021validation}. Moreover, recovering the 3D postures from 2D OpenPose postures without additional sensory information would not be accurate because of the redundancy problem.

To benefit from both sensory information from the robot and OpenPose, we propose to combine them as a multi-sensory posture estimation system to provide an accurate and occlusion-robust estimation in teleoperation that can also handle redundancy problem.We formalize posture estimation as a probabilistic inference problem, in which we measure the leader robot's trajectory~(pose and velocity) and the 2D position of anatomical landmarks of the body from OpenPose in the camera coordinate frame as the observation, and infer the unobserved 3D human posture (joint angles and angular velocities). The combination of a simple RGB camera and the leader robot (which is already available from the teleoperation with no extra cost) would still be low cost compared to MoCap systems, depth sensors, or multiple camera systems.

We compare the combined method with posture estimation solely from (1) robot's trajectory and (2) 2D postures from OpenPose. Similar to our previous work, we use the \textit{circle point analysis~(CPA)} for segment length estimation of the human body as we showed that it has higher accuracy compared to other approaches~\cite{yazdani2020leader}. We also impose physical limits on joint angles and check the validity of the posture based on the posture-dependant ranges of human motion provided by~\cite{jiang2018data}. We incorporate multiple observations over time enabling us to perform inference using a standard particle filter. Then, we use the estimated posture over the course of the task to assess the user's risk of WMSDs using RULA~\cite{McAtamney1993rula}, a standard measure in the ergonomics and safety community. Moreover, we conduct a human subject study and compare the posture estimation results from the different approaches mentioned above, and the postures from the gold-standard Optitrack\footnote{\url{https://optitrack.com/}} MoCap system. Finally, we show that our posture estimation approaches has good accuracy in risk assessment, comparing to the postures from a MoCap system.\

We summarize the contributions of this paper as the following:
\begin{itemize}
\item Modeling the 3D posture estimation in teleoperation using sensory information from the leader robot's trajectory and 2D postures from OpenPose as a partially-observable dynamic system.
\item Infer the 3D posture from the observations using a particle filter,
\item Conducting a human subject study to evaluate and compare the performance of each approach in estimating the 3D posture and ergonomics assessment in teleoperation.
\end{itemize}

We structure the remainder of this paper as follows. We fist review the related literature in the field and focus on the recent state-of-the-art work. We then formalize the 3D posture estimation using three types of observations: solely from the robot, solely from OpenPose, and multi-sensory observation from the robot and OpenPose. Later, we introduce our approach to solve the partially-observable posture estimation problem and provide details on our human subject study. To conclude, we present and discuss the results of our experiment.

Dataset, code, and supplementary materials are available at \href{https://sites.google.com/view/posture-estimation-in-teleop/home}{https://sites.google.com/view/posture-estimation-in-teleop/robot-openpose}

\section{Related Work}
\label{sec:related-work}
In this section, we provide an overviews of the literature on posture estimation in human-robot interaction and it the key challenges and components of them.
\subsection{Posture Estimation in Human-Robot Interaction}
Human posture estimation has been an important tool in various areas of human-robot interaction such as socially-assistive robots~\cite{shi2021toward,spitale2022socially}, pHRI~\cite{figueredo2021planning, vianello2021human}, teleoperation~\cite{lin2022intuitive,zhou2016comparative, dragan2013teleoperation}, engagement analysis~\cite{hadfield2019deep} human-aware decision making and planning~\cite{chalvatzaki2019learn}, and situation awareness~\cite{dini2017measurement}. 
Posture estimation in pHRI has been used mainly as a tool to derive other metrics for HRI evaluation. For instance, researchers have discussed ergonomics assessment and safety-based factors such as joint overloading~\cite{kim2018ergotac, fang2018real}, the separation distance between the robot and human~\cite{Zanchettin2016safety}, and muscular fatigue~\cite{peternel2018online}. In teleoperation estimated posture has been used to directly control a remote robot~\cite{vartholomeos2016design,buzzi2018uncontrolled,martinez}. The idea of estimating the 3D posture of a human teleoperator only from the trajectory of the leader robot was introduced concurrently in our previous work~\cite{yazdani2020leader} and by Rahal et al. in~\cite{rahal2020caring}. Rahal solved the IK of the 7-DOF human arm holding the stylus of a leader robot in teleoperation, assuming the known position of human's shoulder. Later, Vianello et al.~\cite{vianello2021human} used the same idea to predict human posture in pHRI given the robot trajectory executed in a collaborative scenario. They formalized the problem as the prediction of the human joints velocity given the current posture and robot end-effector velocity over demonstrated human movements.

\subsection{Redundancy Resolution of the Human Skeletal Model}
The inverse kinematics problem of human kinematics has infinite solutions due to the redundancy of the kinematic chain. General methods for inverse kinematics of redundant systems use the pseudo-inverse of the Jacobian matrix to minimize the error of the end-effector using e.g. the Newton-Raphson method~\cite{lynch2017modern}. Several studies have focused on the improvement of the solution of the problem. For instance, Kim et al. formulated the inverse kinematics problem byu defining the swivel angle, the rotation angle of the plane defined by the upper and lower arm around a virtual axis that connects the shoulder and wrist joints. They analyzed reaching tasks recorded with a motion capture system indicating that the swivel angle is selected such that when the elbow joint is flexed, the palm points to the head, and they used this criterion to resolve the human arm redundancy~\cite{kim2012redundancy}. Rahal et al~\cite{rahal2020caring} used the same formulation of swivel angle and used a heuristic that minimizes the deviation from neutral posture to resolve the redundancy. The above approaches for redundancy resolution do not always hold across different tasks, and therefor cannot be generalized to all applications. Another researcher tried to include the intrinsic principles of human arm motion including both skeletal kinematics and muscle strength properties~\cite{li2018efficient}. Although their approach results in bio-mimetic arm postures, the relationship between elements of joint torques in their skeletal model needs more investigation and their muscle effort in their model is realistic only for static postures.

Recently, learning-based approaches have been used for resolving the redundancy of human arm and generating human-like motions~\cite{yang2021humanoid,kim2010learning,yang2018dmps, vianello2021human}. For instance, authors in~\cite{vianello2021human} learned the distribution of the null space of the Jacobian and the weights of the weighted pseudo-inverse from demonstrated human movements. In~\cite{deng2022human} authors designed one-class support vector machine model to classify human-like postures. Then, they used the redundancy characteristic of a 7-DOF robotic arm with a linear regression model to enhance the search of human-like postures.  

In all of the above literature, the redundancy of the 7-DOF human arm model was the target, however, in our posture estimation in teleoperation, we consider the redundancy of the full 10-DOF upper-extremity of human skeleton including torso and arm. Although going deep in the redundancy resolution of human arm models is not the in the scope our work, in our previous work~\cite{yazdani2020leader} we used the prior information about the task including the torso being almost vertical and initiating the neutral posture to resolve the redundancy. Here in this paper, we explore the effect of using low-cost multiple sensory information to disambiguate the redundancy of the human skeletal model.

\subsection{Human Joint Range of Motion}
Another challenge in human posture estimation is defining the joint ranges of motion. The literature shows that the range of motion for a joint varies depending on the positions of other joints~(inter-joint dependency) or other degrees-of-freedom in the same joint~(intra-joint dependency), and vary by gender and person~\cite{wang1998three,jiang2018data}. Authors in~\cite{akhter2015pose} developed a discontinuous mathematical model for posture-dependant range of motion based on recorded MoCap of human motions and checked the validity of a full-body posture. Later, Jiang et al.~\cite{jiang2018data} used the above model to label the validity of a set of randomly-generated postures and learned a differentiable neural network based on the generated data and used it as a constraint in the inverse kinematics optimization. However, their arm model is limited to shoulder and elbow and it does not include the human wrist. In our work, we use their learned model on top of the standard human range of motion from biomechanics~\cite{levangie2000joint,nasa}.

\subsection{Postural Assessment in Human-Robot Interaction}
Ergonomics risk assessment of tasks and workplaces is an important part of the process to reduce the number of WMSDs~\cite{ismail2010evaluation,khodabakhshi2014ergonomic}. Ergonomists have provided simple and easy to calculate risk assessment tools which are common in practice. NASA TLX~\cite{hart1988development}, RULA~(rapid upper limb assessment)~\cite{McAtamney1993rula}, REBA~(rapid entire body assessment)~\cite{hignett2000rapid}, strain index~\cite{steven1995strain}, and ACGIH TLV~\cite{Kapellusch2014strain} are among the common risk assessment tools in ergonomics. Importantly, these models are supported by extensive human subject studies that validate their effectiveness on reducing ergonomic risk factors~\cite{khodabakhshi2014ergonomic}. pHRI researchers have also proposed computational models for human comfort and ergonomics, including peripersonal space~\cite{chen2018planning}, muscle fatigue~\cite{peternel2017towards}, joint overloading~\cite{kim2017anticipatory}, and muscular comfort~\cite{chen2018planning}. RULA and REBA are among the most popular risk assessment tools in the ergonomics community which both rely mostly on the human posture~(i.e. joint angles) and provide quantitative scores. The inputs to the both models are the joint angles of the human, and they output discrete, integer scores where the higher number indicates the higher risk of WMSDs, i.e. lower human comfort~\cite{figueredo2020human}. Moreover, there are interpretations for different ranges of RULA scores which provide the final assessment results. We show some interpretations in Fig.~\ref{fig:rula_subject}. Those features make RULA the most suitable assessment tool to assess the upper extremity tasks that are common in teleoperation and pHRI. We also proposed DULA and DEBA, the differentiable and continuous versions of RULA and REBA risk assessment tools in our previous work~\cite{yazdani2022dula} to be used in gradient-based postural optimization in pHRI as a part of our ergonomically intelligent system framework. We use the generic RULA for assessing posture in this paper. 

%%%%%%%%%%%%%%%%%%%%%%%%%%%%%%%%%%%%%%%%%%%%%%%%%%%%%%%%%%%%%%%%%%%%%%%%%%%%%%%%%%
%%%%%%%%%%%%%%%%%%%%%%%%%%%%%%%%%%%%%%%%%%%%%%%%%%%%%%%%%%%%%%%%%%%%%%%%%%%%%%%%%%
%%%%%%%%%%%%%%%%%%%%%%%%%%%%%%%%%%%%%%%%%%%%%%%%%%%%%%%%%%%%%%%%%%%%%%%%%%%%%%%%%%
\section{Problem Statement}
\label{sec:problem_statement}
Estimating the human posture includes two main parts: (1) estimating the segment lengths, and (2) estimating the joint angles. In this section, we formulate these problems based on different types of observations, and define their corresponding parameters. First, we define the kinematics of human skeletal model. Then, we formulate the 3D posture estimation from different types of sensory information. Finally, we define the problem of segment lengths estimation.

\subsection{Human kinematics model}
\label{sec:approach-kinematic}
\begin{figure}[t!]
% \center
\includegraphics[width = 0.4\textwidth]{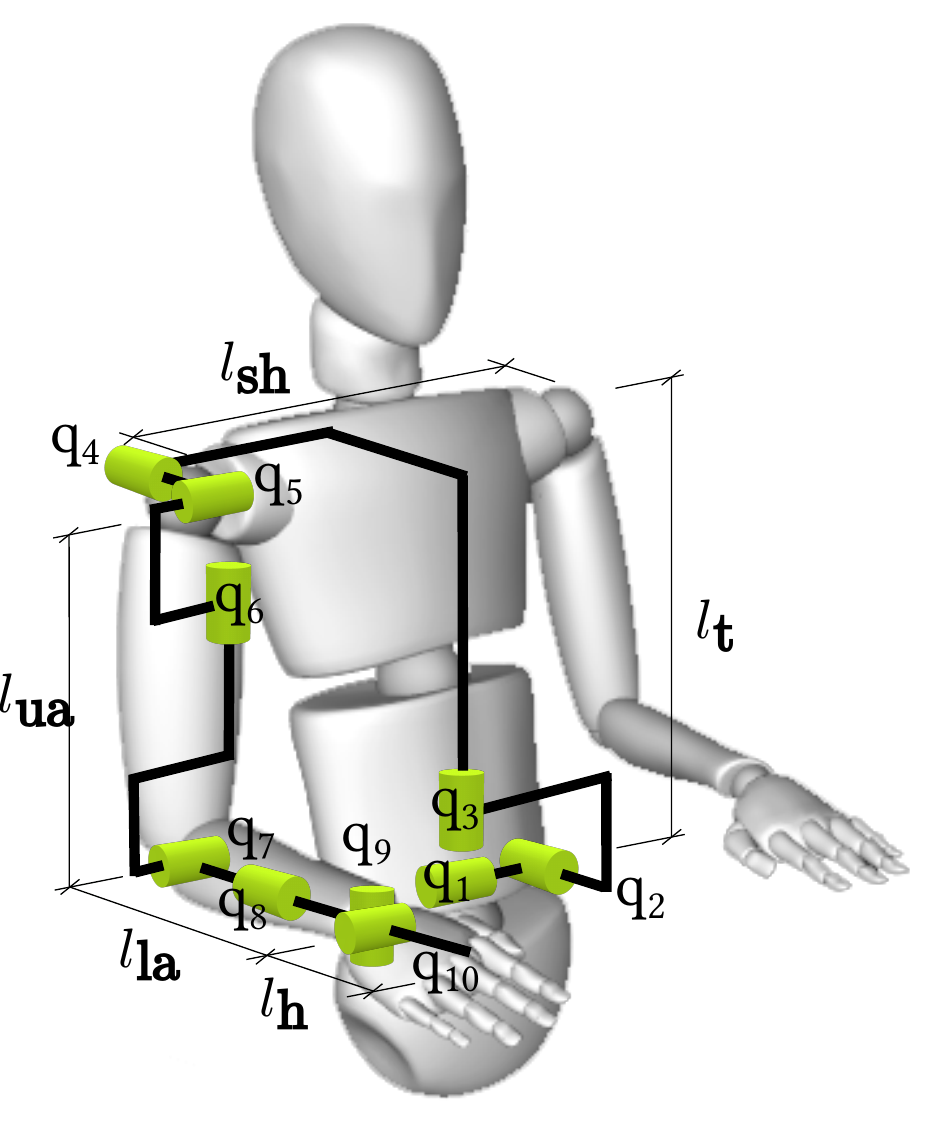}
\caption{Kinematics model of the 10-DOF human upper body, adapted from ~\cite{yazdani2020leader}}
\label{fig:human_model}
\end{figure}

We model the kinematics of a human sitting on a chair and interacting with the leader robot with a 10-DOF kinematics chain, starting from the torso and ending with the right hand as shown in Fig.~\ref{fig:human_model}. We assume that the chair is stationary with a known relative position to the robot, and the human sits on the center of the chair. 

We use the circle point analysis~(CPA) from our previous work~\cite{yazdani2020leader} to estimate the segment lengths. 
We define the human's \textit{state variables} as $\mathbf{q}=[q^{j}]_{j =1:10}$, $\dot{\mathbf{q}}=[\dot{q}^{j}]_{j =1:10}$ where $q^j$ represents the angle of joint $j$ (shown in Fig.~\ref{fig:human_model}). We assume that the user's hand stays attached to the leader robot's stylus, as such we can use a fixed transformation matrix to transfer the pose of the hand from the human's frame to the robot's frame.

From the kinematics of human motion, we find joint angles and velocities based on the previous step from function $f$ as follows:
\begin{equation}
    \begin{bmatrix}
    \mathbf{q}_t\\
    \dot{\mathbf{q}}_t
    \end{bmatrix} = f([\mathbf{q}_{t-1};\dot{\mathbf{q}}_{t-1};\ddot{\mathbf{q}}_{t-1}]) = \begin{bmatrix}
    1&\Delta t\\
    0&1
    \end{bmatrix} \begin{bmatrix}
    \mathbf{q}_{t-1}\\
   \dot{\mathbf{q}}_{t-1}
    \end{bmatrix}+\begin{bmatrix}
    0\\
    \ddot{\mathbf{q}}_{t-1}\Delta t 
    \end{bmatrix}
\label{eq:kinematic_update}    
\end{equation}
We model joint accelerations generated from a Gaussian distribution $\ddot{\mathbf{q}}_{t-1}\sim\mathcal{N}(0,\Tilde{\mathbf{\Sigma}}_v)$. Since $dt$ is fixed, setting \(\mathbf{\Sigma}_v = \Tilde{\mathbf{\Sigma}}_v\cdot dt\) results in:
\begin{equation}
p(\dot{\mathbf{q}}_t\mid \dot{\mathbf{q}}_{t-1})\sim \mathcal{N}(\dot{\mathbf{q}}_{t-1},\mathbf{\Sigma}_v)
\label{eq:vel_update}
\end{equation}

\subsection{3D Posture Estimation Solely from OpenPose 2D Postures}
At each time step, OpenPose provides 2D position of the keypoints corresponding to anatomical landmarks on human body in camera coordinate frame. To recover the 3D posture, we seek to find the 3D posture trajectory $\mathbf{\tau}=[\mathbf{q}_t;\dot{\mathbf{q}}_t]_{t=1:T}$ that its projection into the 2D camera frame matches the posture from OpenPose $(\mathbf{u},\mathbf{v})$ and obeys the human motion model during the task. We can formulate the 3D posture estimation from 2D OpenPose postures as 
\begin{align}
\mathbf{\tau}^*=&\argmin_{\mathbf{\tau}}\sum_{t=1}^T(\sum_{m=1}^M\!\begin{aligned}[t]
&||\mathcal{P}\cdot \phi(\mathbf{q}^{1:m}_t, \mathbf{\psi}) - (u^m,v^m)||_{\Sigma_1}^2)+\\&||\left[\mathbf{q}_{t};\dot{\mathbf{q}}_{t}\right] - f([\mathbf{q}_{t-1};\dot{\mathbf{q}}_{t-1};\ddot{\mathbf{q}}_{t-1}])||_{\Sigma_2}^2, \label{eq:problem_statement_openpose}\end{aligned} \\
                & s.t. \qquad \mathbf{q}_{\mathrm{min}}^j \leq \mathbf{q}^j \leq \mathbf{q}_{\mathrm{max}}^j \qquad \mathrm{for~} j \mathrm{~in~} [1, 2,~\dots, \mathrm{~number~of~joints}]\nonumber
\end{align}
where $\psi = [l_{\mathrm{torso}}, l_{\mathrm{shoulder}}, l_{\mathrm{upper~arm}}, l_{\mathrm{lower~arm}}, l_{\mathrm{hand}} ]$ are the segment lengths depicted in Fig.~\ref{fig:human_model}, $\phi(\mathbf{q}_t,\psi)$ is the forward kinematics of human based up to the grasping point at the posture $\mathbf{q}$, $f$ is the human motion model (see Eq.~\ref{eq:kinematic_update} for more details), $\Sigma_1$ is the Gaussian covariance for keypoint noise in the image space, $\Sigma_2$ is the covariance matrix for joint angles and their angular velocities, and $\mathbf{q}^j_{\mathrm{min}}$ and $\mathbf{q}^j_{\mathrm{max}}$ are the joint limits for joint $j$. We encode the human range of motion as truncated fixed limits on the joint angles based on the biomechanics literature~\cite{levangie2000joint, nasa}. Moreover, we use the learned and posture-dependant joint limit model from \cite{jiang2018data} to ensure the validity of the posture. Unlike our previous work, here we use the full range of motion from the biomechanics literature and we do not narrow the torso range of motion. 

If occlusion occurs in the scene, OpenPose still tries to provides the estimated position of occluded keypoint based on the position of other keypoints, unless it is not possible. Figure~\ref{fig:openpose_wrong_wm} visualizes one of those moments where the the wrist and upper lower arm are occluded. These wrong estimated 2D keypoints will result in high error in the estimated 3D posture.

\subsection{3D Posture Estimation Solely from Robot's Trajectory}
When the leader robot~(or the interacting robot in general pHRI) is the only sensor for posture estimation, we seek to solve the problem of estimating the human joint-space trajectory in teleoperation using only the observed task-space poses and velocities of the leader robot. Another piece of observation that is available by default at each time step is the estimated velocity at the previous time step. We model the physical interaction between the human and the leader robot as an interaction point where the human kinematic chain makes contact with the robot's stylus (Fig.~\ref{fig:setup}).

At each time step, the robot provides an observation as a task-space pose $\mathbf{z}$ and velocity $\dot{\mathbf{z}}$ of the stylus at the interaction point,
\begin{equation}
    [\mathbf{z}_t;\dot{\mathbf{z}}_t]=h(\phi([\mathbf{q}_t;\dot{\mathbf{q}}_t], \mathbf{\psi}))
\end{equation}
where $h$ is the observation function including the transfer of human pose into the corresponding robot stylus pose in the robot's coordinate frame. This defines only a partial observation of the human posture, because of redundancy in the human kinematics and a noisy measurement at the interaction point, which may change slightly during a task.

We seek to estimate $\mathbf{\tau}=[\mathbf{q}_t;\dot{\mathbf{q}}_t]_{t=1:T}$ given the stylus trajectory $\mathcal{Z}=[\mathbf{z}_t;\dot{\mathbf{z}}_t]_{t=1:T}$ that predicts a stylus pose closest to the observed stylus pose and obeys the human motion model $f$~(Eq.~\ref{eq:kinematic_update}):
\begin{align}
\mathbf{\tau}^*=&\argmin_{\mathbf{\tau}}\sum_{t=1}^T\!\begin{aligned}[t]
&||\phi([\mathbf{q}_t;\dot{\mathbf{q}}_t], \mathbf{\psi})-[\mathbf{z}_t;\dot{\mathbf{z}}_t]||_{\Sigma_3}^2+\\&||\left[\mathbf{q}_{t};\dot{\mathbf{q}}_{t}\right] - f([\mathbf{q}_{t-1};\dot{\mathbf{q}}_{t-1};\ddot{\mathbf{q}}_{t-1}])||_{\Sigma_2}^2+\\
&||\dot{\mathbf{q}}_t-\dot{\mathbf{q}}_{t-1}||^2_{\Sigma_4}, \label{eq:problem_statement_robot}\end{aligned} \\
                & s.t. \qquad \mathbf{q}_{\mathrm{min}}^j \leq \mathbf{q}^j \leq \mathbf{q}_{\mathrm{max}}^j \qquad \mathrm{for~} j \mathrm{~in~} [1,2,~\dots, \mathrm{~number~of~joints}]\nonumber
\end{align}
where $\Sigma_3$ is the covariance matrix for position and orientation elements, and $\Sigma_4$ is thecovariance matrix for joint angular velocity elements. The high degree-of-freedom in human kinematics makes this problem a \emph{redundant} problem with an infinite number of solutions. We seek the solution closest to the true posture of the human teleoperator.

\subsection{Multi-Sensory 3D Posture Estimation from OpenPose and Robot's Trajectory}
The multi-sensory posture estimation from a single RGB camera and the robot can be modeled as the combination of the above formulations:
\begin{align}
\mathbf{\tau}^*=&\argmin_{\mathbf{\tau}}\sum_{t=1}^T\!\begin{aligned}[t]
&||\phi([\mathbf{q}_t;\dot{\mathbf{q}}_t], \mathbf{\psi})-[\mathbf{z}_t;\dot{\mathbf{z}}_t]||_{\Sigma_1}^2+\\
&\sum_{m=1}^M(||\mathcal{P}\cdot \phi(\mathbf{q}^{1:m}_t, \mathbf{\psi}) - (u^m,v^m)||_{\Sigma_3}^2)+\\
&||\left[\mathbf{q}_{t};\dot{\mathbf{q}}_{t}\right] - f([\mathbf{q}_{t-1};\dot{\mathbf{q}}_{t-1};\ddot{\mathbf{q}}_{t-1}])||_{\Sigma_2}^2+\\&||\dot{\mathbf{q}}_t-\dot{\mathbf{q}}_{t-1}||^2_{\Sigma_4}, \label{eq:problem_statement_robot_openpose}\end{aligned} \\
                & s.t. \qquad \mathbf{q}_{\mathrm{min}}^j \leq \mathbf{q}^j \leq \mathbf{q}_{\mathrm{max}}^j \qquad \mathrm{for~} j \mathrm{~in~} [1,2,~\dots, \mathrm{~number~of~joints}]\nonumber
\end{align}
All of the above problems are \textit{redundant} problems which in general have an infinite number of solution. Our goal is to find the solutions that are closest to the true posture of the human.

\subsection{Segment Length Estimation}
Segment length estimation using the vision-based 3D posture estimation systems is straightforward: each segment length is calculated based on the 3D position of the tracked markers at the each end of the segment. However, this process still suffers from inaccuracy due to imperfect marker placement and skin artifacts in MoCap systems~\cite{metcalf2020quantifying} and model, sensor, and marker tracking uncertainties in markerless approaches~\cite{d2021validation}.

We can define the estimation of segments lengths solely from the robot as 
\begin{equation}
    \psi^* = \argmin_\psi \sum_{t=1}^T |\phi(\mathbf{q}_t,\psi) - \mathbf{z}_t|^2,
    \label{eq:seg_len_robot}
\end{equation}

When the 3D posture from OpenPose is the only sensory information, we can formulate the segment length estimation as
\begin{equation}
    \psi^* = \argmin_\psi \sum_{t=1}^T \sum_{m=1}^M |\mathcal{P}\cdot  \phi(\mathbf{q}^{1:m}_t,\psi)  -(u^m,v^m) |^2,
    \label{eq:seg_len_openpose}
\end{equation}
where $M$ is the number of related keypoints from OpenPose, $\mathbf{q}^{1:m}_t$ is the set of joint angles contributing to the pose of the keypoint $m$, and $\mathcal{P}$ is the projection matrix of the camera that transfers 3D positions of keypoints in the world frame into 2D pixels $(u,v)$ in the camera frame.

Solving the above optimizations are challenging and cumbersome when the posture $\mathbf{q}$ is unknown, hence usually they are solved for segment lengths and joint angles simultaneously. 
For the following parts of this section, we assume that the segments lengths are known. 
%%%%%%%%%%%%%%%%%%%%%%%%%%%%%%%%%%%%%%%%%%%%%%%%%%%%%%%%%%%%%%%%%%%%%%%%%%%%%%%%%%
%%%%%%%%%%%%%%%%%%%%%%%%%%%%%%%%%%%%%%%%%%%%%%%%%%%%%%%%%%%%%%%%%%%%%%%%%%%%%%%%%%
%%%%%%%%%%%%%%%%%%%%%%%%%%%%%%%%%%%%%%%%%%%%%%%%%%%%%%%%%%%%%%%%%%%%%%%%%%%%%%%%%%
\section{Approach}
\label{sec:approach}
In this section, we introduce approximate solutions for Eqs.~\ref{eq:problem_statement_openpose},~\ref{eq:problem_statement_robot},~and \ref{eq:problem_statement_robot_openpose} by solving the partially observable posture estimation problem using a particle filter. Initially, we model the observation from the sensors. Later, we discuss adopting a particle filter for inference in posture estimation.

%%%%%%%%%%%%%%%%%%%%%%%%%%%%%%%%
\subsection{Observation Likelihood}
When we use the robot trajectory as the observation, we model the observation likelihood function by a Gaussian distribution over the hand's pose and velocity as the end-effector of the human kinematic chain:
\begin{equation}
p([\mathbf{z}_t,\dot{\mathbf{z}}_t]\mid [\mathbf{q}_{t},\dot{\mathbf{q}}_{t}]) =\mathcal{N}(\phi(\mathbf{q}_t,\dot{\mathbf{q}}_t,\mathbf{\psi}),\mathbf{\Sigma}_{K})
\label{eq:weight_robot}
\end{equation}
in which $\mathbf{\Sigma}_K$ is the kinematic covariance matrix. 

When using the 2D posture from OpenPose as the observation, we define the observation likelihood function by a Gaussian distribution over the 2D projected position of each keypoint:
\begin{equation}
p([u^m_t; v^m_t]\mid [\mathbf{q}_{t};\dot{\mathbf{q}}_{t}]) =\mathcal{N}(\mathcal{P}\cdot  \phi(\mathbf{q}^{1:m}_t,\mathbf{\psi}),\mathbf{\Sigma}_{P})
\label{eq:weight_openpose}
\end{equation}
where $\mathbf{\Sigma}_{P}$ is the projection covariance matrix.

When combining the two observations as a multi-sensory posture estimation, then 
\begin{equation}
p(\begin{bmatrix}
    \begin{bmatrix}
    u^1_t\\v^1_t
    \end{bmatrix}\\\vdots\\\begin{bmatrix}
    u^M_t\\v^M_t
    \end{bmatrix}\\\begin{bmatrix}
    \mathbf{z}_t\\\dot{\mathbf{z}_t}
    \end{bmatrix}
    \end{bmatrix}\mid \begin{bmatrix}
    \mathbf{q}_{t}\\\dot{\mathbf{q}}_{t} \end{bmatrix}) =\mathcal{N}(\begin{bmatrix}
    \mathcal{P}\cdot  \phi(\mathbf{q}^{1:1}_t,\psi)\\ \vdots \\ \mathcal{P}\cdot  \phi(\mathbf{q}^{1:M}_t,\psi)\\\phi(\mathbf{q}_t,\dot{\mathbf{q}}_t,\mathbf{\psi})
    \end{bmatrix},\begin{bmatrix}
    \mathbf{\Sigma}_{P}\\ \vdots \\ \mathbf{\Sigma}_{P}\\\mathbf{\Sigma}_{K}
    \end{bmatrix})
\label{eq:weight_robot_openpose}
\end{equation}

%%%%%%%%%%%%%%%%%%%%%%%%%%%%%%%
\subsection{Particle Filter for Posture Estimation}
\label{sec:approach-pf}
We approximate the solution for the partially observable posture estimation problems in Section~\ref{sec:problem_statement} by using a particle filter ~\cite{thrun2005probabilistic}. We initialize particles from a uniform distribution over the joint range of motion and set the initial angular velocities to zero. For the rest of the paper, the superscript in the bracket over the parameters represents the corresponding particle number.
\begin{equation}
{\mathbf{q}}_0^{[n]}\sim \mathcal{U}(\mathbf{q}_{\mathrm{min}},\mathbf{q}_{\mathrm{max}}),\quad \dot{\mathbf{q}}_{0}^{[n]}=0 \qquad n=1,...,N, \label{eq:process}
\end{equation}
where $N$ is the number of particles.

Each particle is propagated in time based on the kinematics of human motion using Eq.~\ref{eq:kinematic_update}. Then, the particles are weighted based on their corresponding observation likelihood functions from Eqs.~\ref{eq:weight_robot},~\ref{eq:weight_openpose}, or~\ref{eq:weight_robot_openpose} defined as the \textit{innovation error} between the estimated pose of the stylus and the observed pose from the leader robot, and/or the projected 2D estimate of keypoints  and observed keypoints from OpenPose. We use the multivariate Gaussian distribution to define the likelihood weighting function for each case: 
\begin{align}
w_t^{[n]}=&v_p\cdot \mathrm{det}(2\pi \mathbf{\Sigma}_{K})^{-\frac{1}{2}} \cdot\exp \{-\frac{1}{2}([\mathbf{z}_t,\dot{\mathbf{z}}_t]-\qquad \mathrm{(Observation:~robot)}\nonumber\\ &\phi(\mathbf{q}_t,\dot{\mathbf{q}}_t,\mathbf{\psi}))^T
\mathbf{\Sigma}_{K}^{-1}([\mathbf{z}_t,\dot{\mathbf{z}}_t]-\phi(\mathbf{q}_t,\dot{\mathbf{q}}_t,\mathbf{\psi}))\}
\label{weight_robot2}
\end{align}

\begin{align}
w_t^{[n]}=&v_p\cdot \mathrm{det}(2\pi \mathbf{\Sigma}_{P})^{-\frac{1}{2}} \cdot\exp \{-\frac{1}{2}(\begin{bmatrix}
    \begin{bmatrix}
    u^1_t\\v^1_t
    \end{bmatrix}\\\vdots\\\begin{bmatrix}
    u^M_t\\v^M_t
    \end{bmatrix}
    \end{bmatrix}- \qquad \mathrm{(Observation:~OpenPose)}\nonumber\\ &\begin{bmatrix}
    \mathcal{P}\cdot  \phi(\mathbf{q}^{1:1}_t,\psi)\\ \vdots \\ \mathcal{P}\cdot  \phi(\mathbf{q}^{1:M}_t,\psi)
    \end{bmatrix} ) ^T\mathbf{\Sigma}_{P}^{-1}(\begin{bmatrix}
    \begin{bmatrix}
    u^1_t\\v^1_t
    \end{bmatrix}\\\vdots\\\begin{bmatrix}
    u^M_t\\v^M_t
    \end{bmatrix}
    \end{bmatrix} - \begin{bmatrix}
    \mathcal{P}\cdot  \phi(\mathbf{q}^{1:1}_t,\psi)\\ \vdots \\ \mathcal{P}\cdot  \phi(\mathbf{q}^{1:M}_t,\psi)
    \end{bmatrix} )\}
\label{weight_openpose2}
\end{align}

\begin{align}
w_t^{[n]}=&v_p\cdot \mathrm{det}(2\pi \begin{bmatrix}
    \mathbf{\Sigma}_{P}\\ \vdots \\ \mathbf{\Sigma}_{P}\\\mathbf{\Sigma}_{K}
    \end{bmatrix})^{-\frac{1}{2}} \cdot\exp \{-\frac{1}{2}(\begin{bmatrix}
    \begin{bmatrix}
    u^1_t\\v^1_t
    \end{bmatrix}\\\vdots\\\begin{bmatrix}
    u^M_t\\v^M_t
    \end{bmatrix}\\\begin{bmatrix}
    \mathbf{z}_t\\\dot{\mathbf{z}_t}
    \end{bmatrix}
    \end{bmatrix}- \qquad \mathrm{(Observation:~robot+OpenPose)}\nonumber\\ &\begin{bmatrix}
    \mathcal{P}\cdot  \phi(\mathbf{q}^{1:1}_t,\psi)\\ \vdots \\ \mathcal{P}\cdot  \phi(\mathbf{q}^{1:M}_t,\psi)\\ \phi(\mathbf{q}_t,\dot{\mathbf{q}}_t,\mathbf{\psi})
    \end{bmatrix} ) ^T\begin{bmatrix}
    \mathbf{\Sigma}_{P}\\ \vdots \\ \mathbf{\Sigma}_{P}\\\mathbf{\Sigma}_{K}
    \end{bmatrix}^{-1}(\begin{bmatrix}
    \begin{bmatrix}
    u^1_t\\v^1_t
    \end{bmatrix}\\\vdots\\\begin{bmatrix}
    u^M_t\\v^M_t
    \end{bmatrix}\\\begin{bmatrix}
    \mathbf{z}_t\\\dot{\mathbf{z}_t}
    \end{bmatrix}
    \end{bmatrix} - \begin{bmatrix}
    \mathcal{P}\cdot  \phi(\mathbf{q}^{1:1}_t,\psi)\\ \vdots \\ \mathcal{P}\cdot  \phi(\mathbf{q}^{1:M}_t,\psi)\\ \phi(\mathbf{q}_t,\dot{\mathbf{q}}_t,\mathbf{\psi})
    \end{bmatrix} )\}
\label{weight_robot_openpose2}
\end{align}
where $v_p \in \mathbb{R}, 0\leq v_p \leq 1$, encodes the validity of the posture as output of the learned neural network from~\cite{jiang2018data}.

When any observation is received, each particles is weighted by the observation likelihood function. After normalizing the weights, we re-sample particles based on their weights; we generate more new particles around the particles with higher weights. This process repeats during the tasks and we report the \textit{most probable particle} as the estimated posture at each time.

%%%%%%%%%%%%%%%%%%%%%%%%%%%%%%%%%%%%%%%%%%%%%%%%%%%%%%%%%%%%%%%%%%%%%%%%%%%%%%%%%%%
%%%%%%%%%%%%%%%%%%%%%%%%%%%%%%%%%%%%%%%%%%%%%%%%%%%%%%%%%%%%%%%%%%%%%%%%%%%%%%%%%%%
\section{Implementation and Experimental Protocol}
\label{sec:implementation}
Here, we provide details on our approach, experimental setup, and human subject study to estimate the 3D posture of a human operator performing teleoperation.
\subsection{Human Subject Study}
\begin{figure}[t!]
\center
\includegraphics[width=\textwidth]{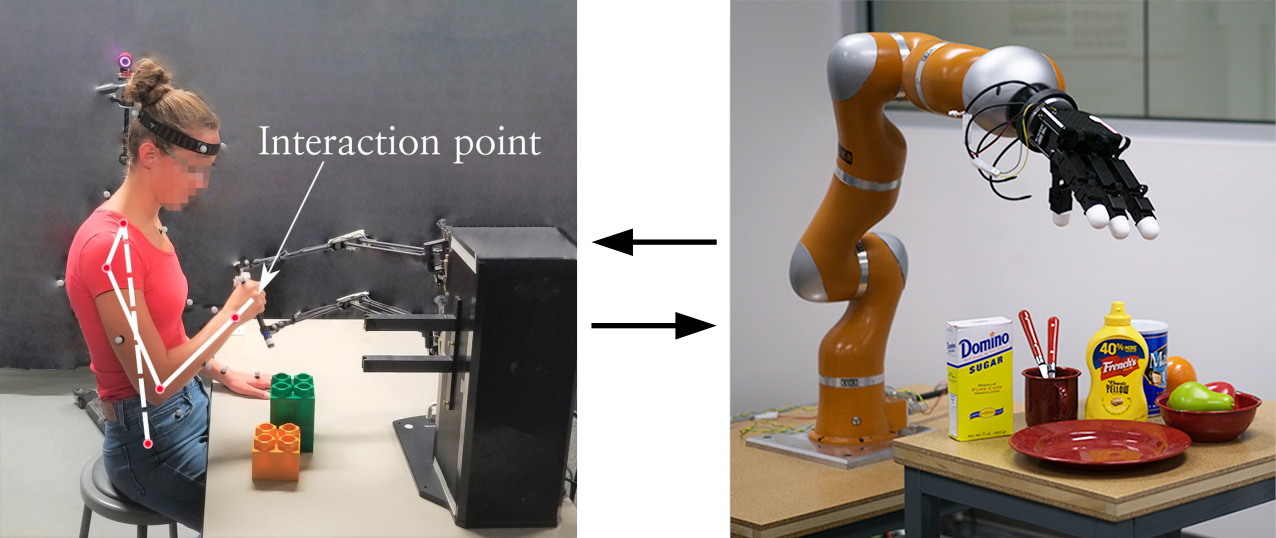}
\caption{Teleoperation setup for human subject experiments including a Quanser HD$^2$ haptic interface, adapted from \cite{yazdani2020leader}}
\label{fig:setup}
\end{figure}

We conducted a human subject experiment in which participants interact with a 6-DOF Quanser HD\(^{2}\) haptic interface\footnote{\url{https://www.quanser.com/products/hd2-high-definition-haptic-device/}} as the leader robot~(see Fig.~\ref{fig:setup}). We recorded their upper body motion using a 12-camera Optitrack MoCap system as the gold standard posture estimation method. We also used a single RGB camera to record videos of the participants and used OpenPose on top of that to generate the 2D posture projected into the camera frame. 
We recruited 8 participants~(4 female, 4 male) with ages ranging from 25 to 33 years and heights in the range of $171\pm21$cm. Participants were graduate students from various programs in the University of Utah. They were informed about the study through a recruiting email sent to graduate students, they volunteered to participate in our study. The study did not include any financial compensation for the participants. Initially, participants reviewed the study documents and signed the consent forms. Then, participants' segment lengths were measured by hand and reflective markers were put on the anatomical landmarks on their bodies by the researchers trained by one of the PIs who is an expert in biomechanics and MoCap systems.

None of the participants had any experience with teleoperation of robots, and each one received a 15-min training with the teleoperation system. Then, to perform the segment length estimation by the CPA method, they followed 5 motion routines discussed in~\cite{yazdani2020leader} while sitting on the chair and holding the robot's stylus in hand.

\begin{figure}[t!]
\center
\includegraphics[width=\linewidth]{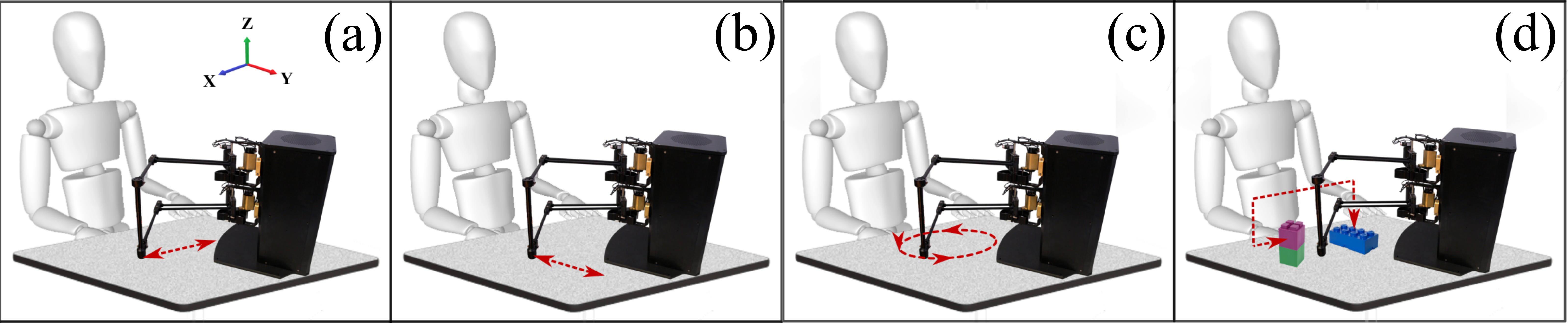}
\caption{The tasks for human subject study: repetitive motions following a straight line in the $X$ direction~(a), a straight line in the $Y$ direction ~(b), a circular path~(c), and repetitive motions between random sides of two blocks positioned at different heights, with an \textit{unprescribed} motion and high range of hand rotation~(d)~\cite{yazdani2020leader}.}
\label{fig:exp_tasks}
\end{figure}

Finally, each participant was asked to performed 4 different tasks visualized in Fig~\ref{fig:exp_tasks}. As the focus of the study is on the interaction between the human and the leader robot, we just provided a printed visual guide on the table in front of the participant for the first three tasks to provide a guideline to follow in order to perform the required motion. However, the participants were not required to follow the path accurately. The participants were not told what posture to initialize the task from and how high they should be above the table to do the task. They chose those based on their comfort. The robot collected data from the participant's motion without exerting any force to their hands.

As neither marker-based nor markerless posture estimation techniques provide ground truth posture, we additionally provide qualitative analysis by overlaying the estimated postures and the MoCap postures on synchronized video frames~(see Fig.~\ref{fig:openpose_wrong_wm}).

Human skeleton used in the MoCap system is much more complicated than the 10-DOF kinematics model we use, specially in torso area. To address this discrepancy, we re-targeted the MoCap posture onto the simple 10-DOF kinematic model by calculating the pose of a virtual torso link from the hip to the neck of the MoCap skeleton. However, as our model uses the fixed segment lengths and the MoCap skeleton uses variable segment lengths. MoCap calculates the segment lengths from the relative position of markers at each time, which are prone to change due to skin artifacts~\cite{metcalf2020quantifying} and imperfect marker placement. Because of this, we see that the skeleton representing the re-targeted MoCap posture does not always follows the human arm in the video frames. So it should be considered when evaluating the qualitative results in Fig.~\ref{fig:openpose_wrong_wm}.

\subsection{OpenPose and Camera Calibration}
\begin{figure}
    \centering
    \begin{minipage}{0.45\textwidth}
            \centering
            \begin{subfigure}{\textwidth}
            \centering
            \includegraphics[width=0.75\textwidth]{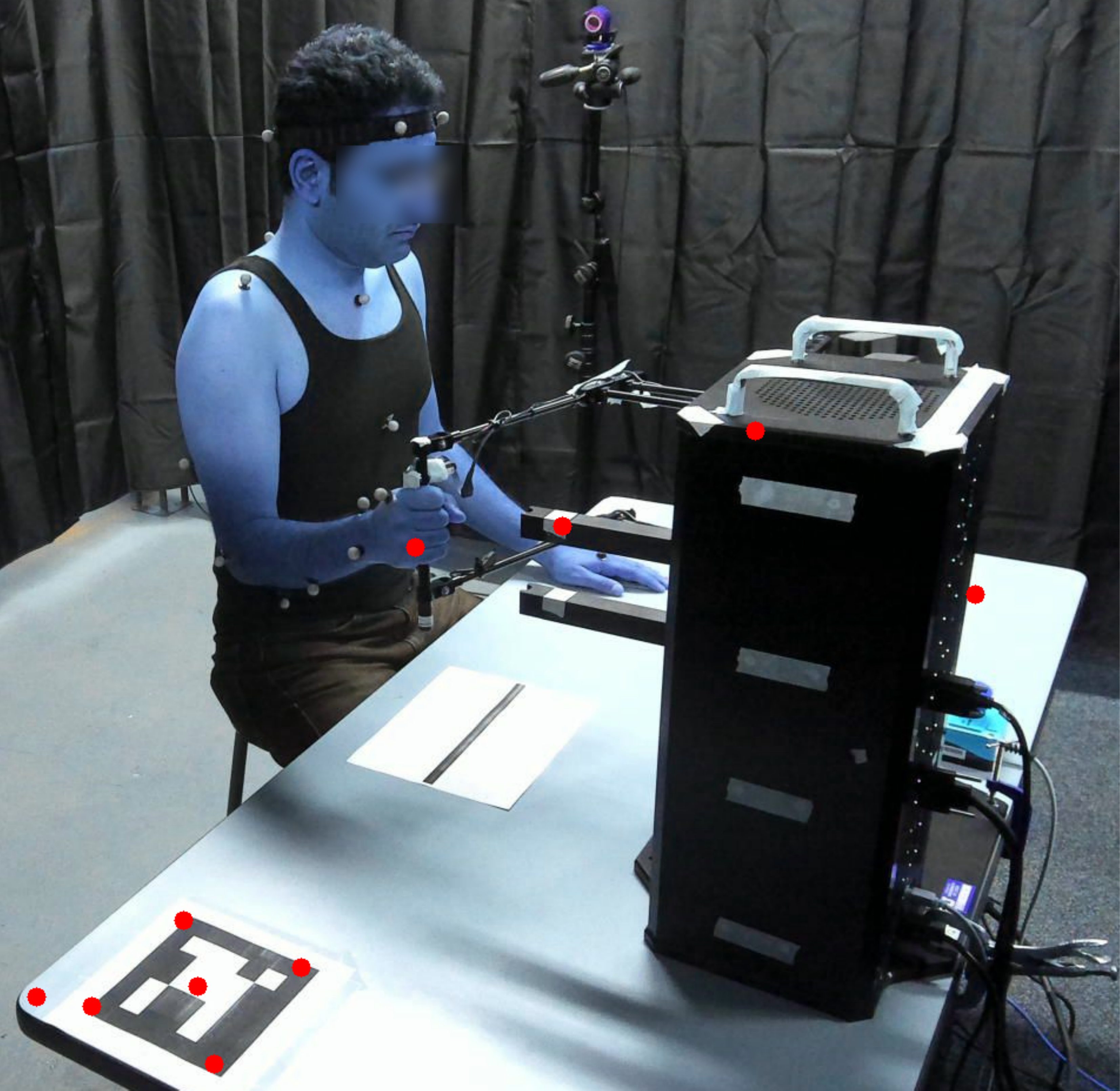}
            \caption{Camera calibration by the ArUco marker.}
            \label{fig:aruco}
            \end{subfigure}
            \\
            \begin{subfigure}{\textwidth}
                % \vspace{0.5mm}
                \centering
                \includegraphics[width=0.75\textwidth]{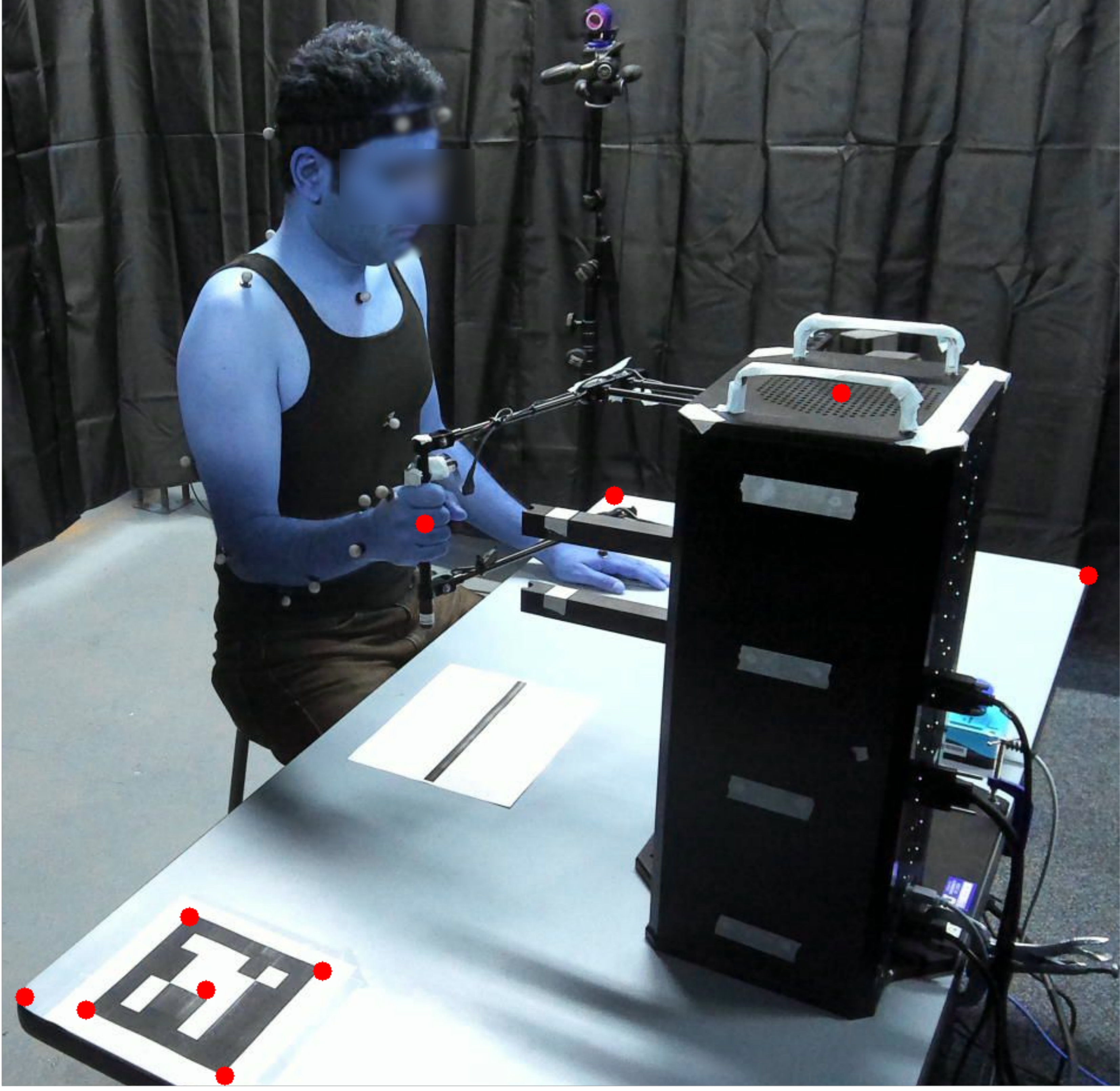}
                \caption{Camera calibration by points with known 3D position.}
                \label{fig:10ots}
            \end{subfigure}

        \caption{Projected points from the scene with known 3D points into the camera image using different calibration methods.}
        \label{fig:camera_calibration}
    \end{minipage}\hfill
    \begin{minipage}{0.45\textwidth}
        \centering
        \vspace{-0.5mm}
        \includegraphics[width=0.92\textwidth]{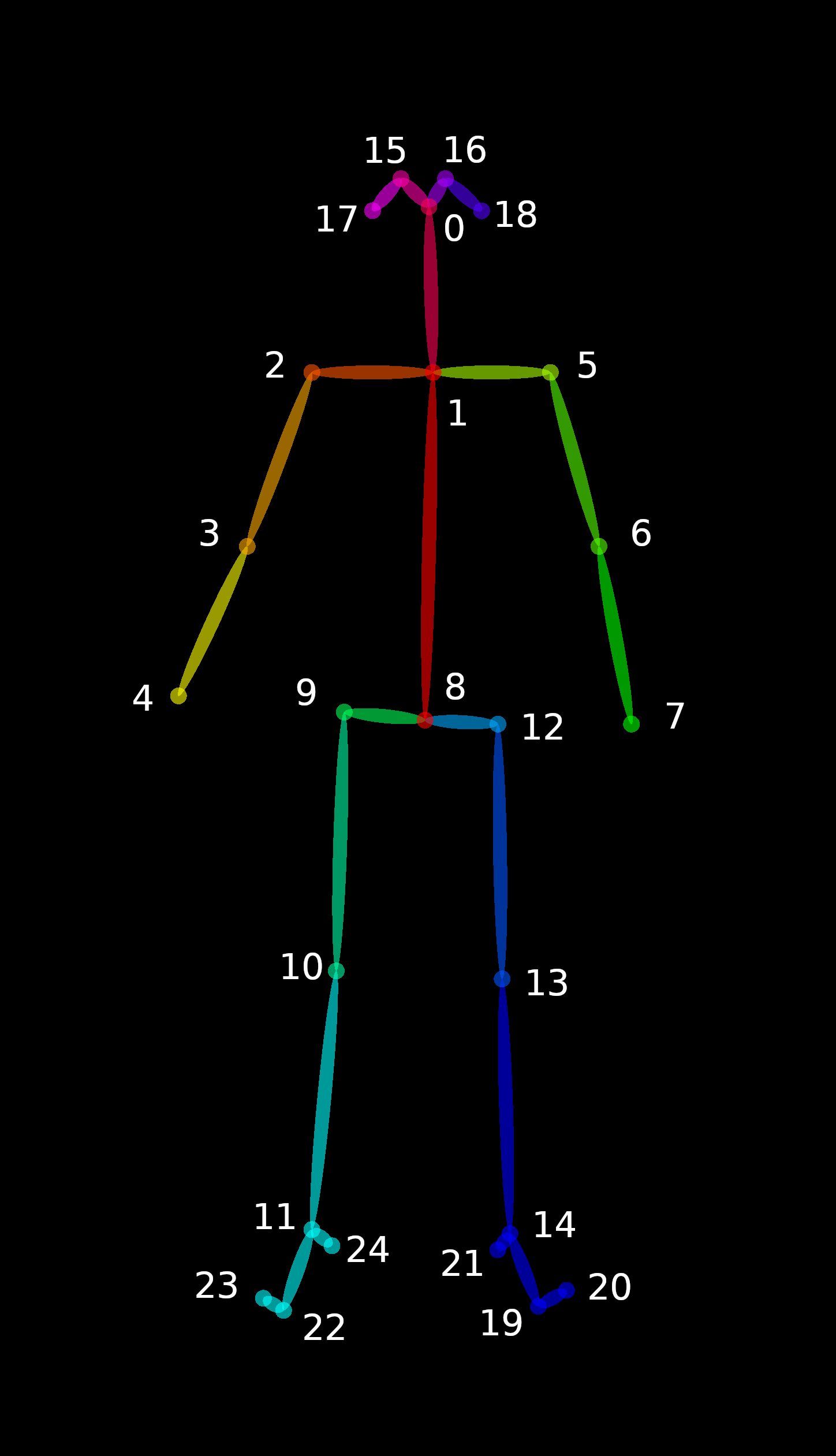} % second figure itself
        \vspace{2mm}
        \caption{Set of keypoints for BODY-25 model in OpenPose~\cite{openpose_docs}.}
        \label{fig:openpose_model}
    \end{minipage}
\end{figure}

Our developed ROS package runs OpenPose in the background and provides the keypoint positions in the camera frame based on the BODY-25 output format shown in Fig.~\ref{fig:openpose_model}. For the purpose of this research, we use keypoint \#1, 2, 3, 4, \& 8 which represent the human torso and the right arm.

Camera calibration is a key part in calculating the weights of particles based on the deviation between the projected anatomical landmarks with the OpenPose keypoints. We calibrate the camera intrinsics using a checkerboard. For extrinsic calibration, we used two approaches: (1) using an ArUco marker attached to the table with known position, and (2) using the known 3D position of 10 points in the scene to calibrate the camera extrinsic parameters based on the common pose estimation method from~\cite{szeliski2010computer}. 
Fig.~\ref{fig:camera_calibration} illustrates the projection of some points in the scene with known 3D positions~(including table corners, robot top corners, center of stylus, and corners on the ArUco marker) into the camera frame using different extrinsic calibration approaches. You can see that the projected points using the known 3D points match better with the original points in the scene, so we use that calibrated extrinsic parameters for our work. Using this we calculated the projection matrix~$\mathcal{P} = P_IT_E$, where $P_I$ and $T_e$ are intrinsic and extrinsic parameters of the camera, respectively. We used the above projection matrix in Sections~\ref{sec:problem_statement} and \ref{sec:approach}.

\begin{figure}[t!]
\center
\includegraphics[width=0.9\linewidth]{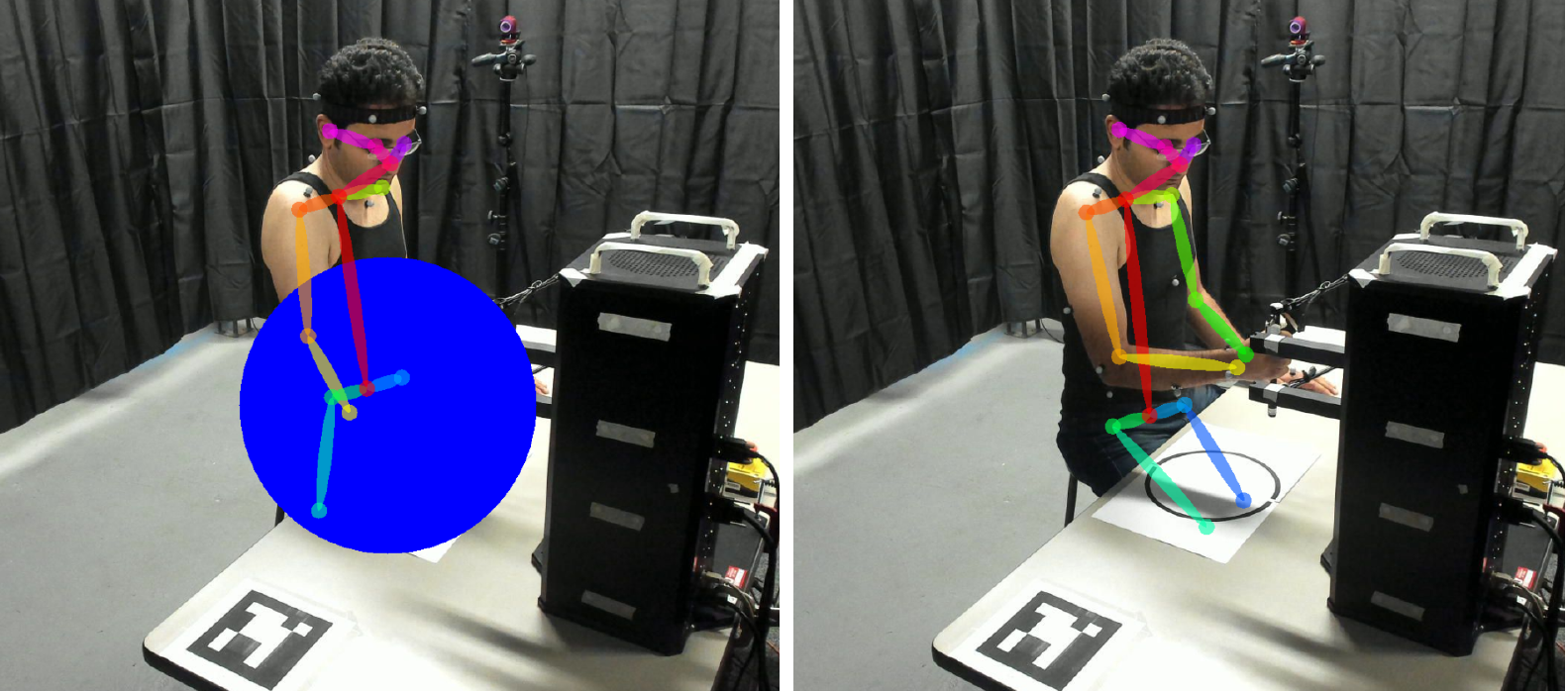}
\caption{To evaluate the robustness to occlusion, we edited the video frames to make artificial occlusion in the video input for OpenPose. You can see the 2D OpenPose posture from the occluded video frame~(left) and non-occluded video frame~(right). }
\label{fig:occlusion_opencv}
\end{figure}

To evaluate the robustness to occlusion for 3D posture estimation solely from OpenPose and our multi-sensory approach, We re-estimate the postures while we occluded some parts of the scene covering using OpenCV in the observation pipeline to edit the video frames fed to the OpenPose. Figure~\ref{fig:occlusion_opencv} visualises the artificial occlusion in the video frames, and its effect on 2D posture estimated by OpenPose.

\subsection{Particle Filter Details}
In our implementation, we use a fixed number of particles (N=500), and used the following parameters:
\begin{equation*}
    \mathbf{\Sigma}_{v} = \texttt{diag}(2, 2, 2, 5, 5, 5, 5, 5, 5, 5)\quad (\frac{deg}{sec})
\end{equation*}
\begin{equation*}
    \mathbf{\Sigma}_{K} = 0.01 \cdot \texttt{diag}(0.01, 0.01, 0.01, 0.01, 0.01, 0.05, 0.1, 0.1, 0.1, 10, 10, 10) \quad ([m,rad,\frac{m}{sec},\frac{rad}{sec}])
\end{equation*}
\begin{equation*}
    \mathbf{\Sigma}_{P} = 0.01 \cdot \texttt{diag}(3, 3, 3, 3, 3) \quad (pixel)
\end{equation*}
We achieved the above numbers by tuning the particle filters on one trial of subject (\#1) performing task (a), and we used the same numbers for all of the other trials.

\subsection{Risk Assessment}
We compute RULA risk assessment scores using the estimated postures from our approach and the MoCap estimates. In calculating the RULA score, we use the following assumptions for all tasks: the human is sitting on a chair, the subject receives minimal intermittent force/load~(less than 2.0Kg), muscle use occurres less than 4x per minute, untwisted and vertical position for neck and torso, and supported legs and feet. We performed all the processes including data collection, synchronization, posture estimation, and ergonomics risk assessment using ROS on a Linux machine with Core-i9 CPU, 32 GB of RAM, and on a NVIDIA 1070Ti GPU.

%%%%%%%%%%%%%%%%%%%%%%%%%%%%%%%%%%%%%%%%%%%%%%%%%%%%%%%%%%%%%%%%%%%%%%%%%%%%%%%%%%%
%%%%%%%%%%%%%%%%%%%%%%%%%%%%%%%%%%%%%%%%%%%%%%%%%%%%%%%%%%%%%%%%%%%%%%%%%%%%%%%%%%%
%%%%%%%%%%%%%%%%%%%%%%%%%%%%%%%%%%%%%%%%%%%%%%%%%%%%%%%%%%%%%%%%%%%%%%%%%%%%%%%%%%%
\section{Results and Discussion}
\label{sec:results}
In this section, we present the results from our human subject experiments. We discuss the performance of the proposed multi-sensory 3D posture estimation approach comparing with the posture estimation solely from the trajectory of the robot, and posture estimation solely from OpenPose. We compare them with the estimated postures from gold-standard a MoCap system and evaluate ergonomics risk assessment using any of the above three approaches. 
%%%%%%%%%%%%%%%%%%%%%%%%%%%%%%%%%%%%%%%%%%%%%%%%%%%%%%%%%
\subsection{Posture Estimation}
\begin{figure}[t]
% \vspace{-3pt}
\begin{centering}
% \hspace{-1cm}
\includegraphics[width=0.9\textwidth]{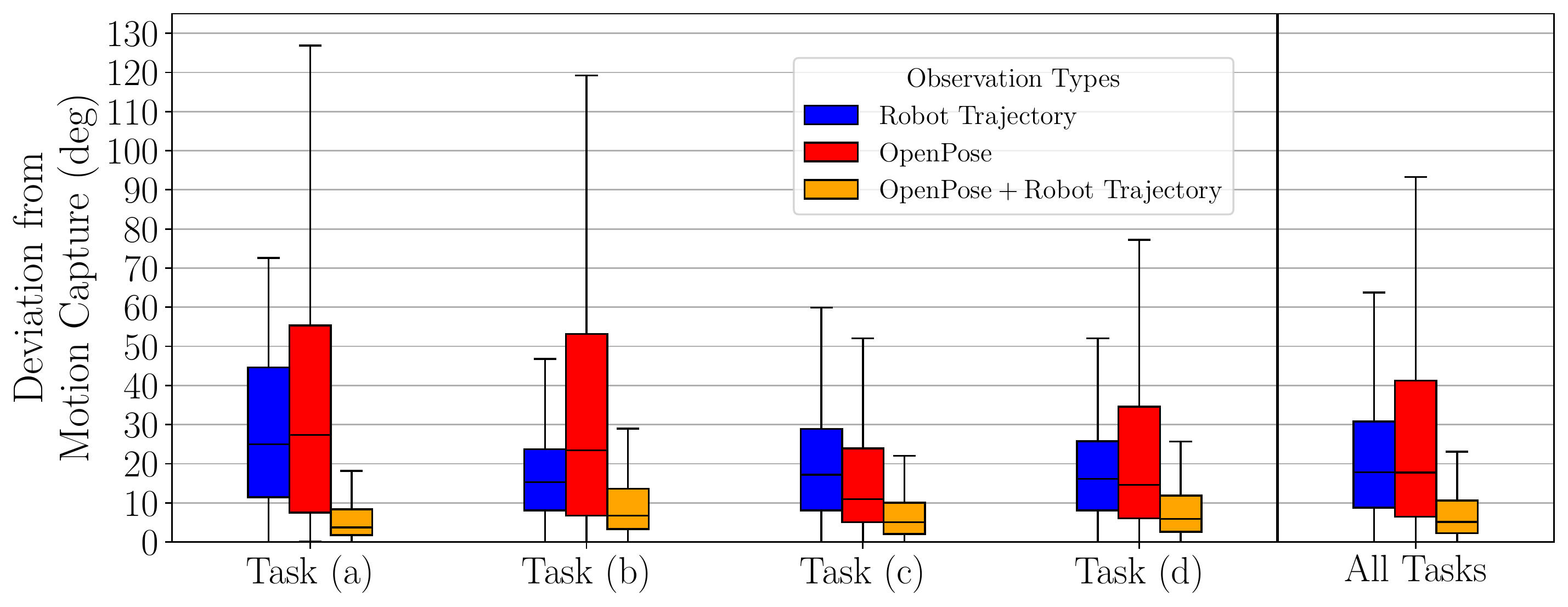}
\caption{Deviation of the estimated postures by three different approaches from MoCap posture among tasks.}
\label{fig:boxplot_tasks}
% \vspace{-0.5cm}
\end{centering}
\end{figure}

Initially we compare the posture estimation accuracy of three approaches by calculating the deviation\footnote{We use the term ``deviation'' instead of ``error'' since the MoCap posture is also an estimate and not ground truth.} of their posture from the MoCap postures. Figure~\ref{fig:boxplot_tasks} presents the deviation for three approaches for all the participants, and trials across our four tasks. As expected, we observe that our multi-sensory approach has significantly lower deviation compared to the other two approaches, and it is true among all of the tasks. The median of deviation for our multi-sensory approach is less than 5 deg with 75\% quantile at 10 deg and upper whisker at 23 deg, which reveals the accuracy of our approach.

\begin{figure}[t!]
% \vspace{-3pt}
\begin{centering}
% \hspace{-1cm}
\includegraphics[width=0.9\textwidth]{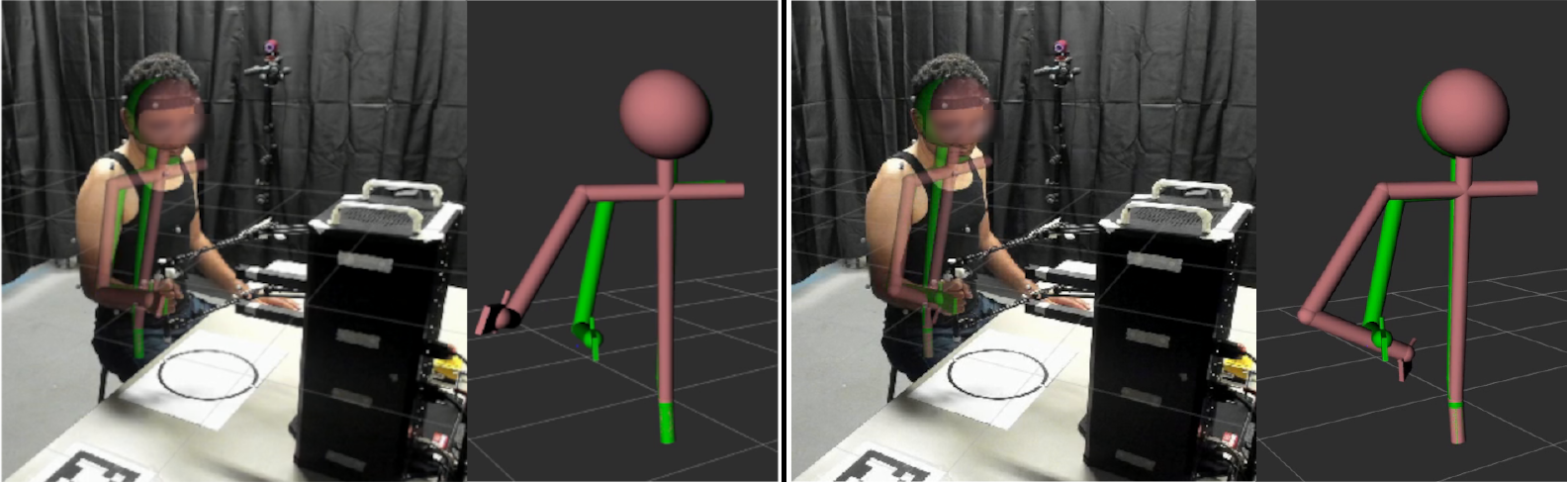}
\caption{Two example of wrong posture estimated by OpenPose. Although the estimated posture~(pink skeleton) matches well with the MoCap posture~(green skeleton) in the camera view, a view from a different angle shows that those postures are deviated highly from each other.}
\label{fig:openpose_wrong_wm}
% \vspace{-0.5cm}
\end{centering}
\end{figure}

Moreover, we see that the deviation for estimated posture solely from the robot trajectory is lower in tasks (b, c, d) compared to task (a). We believe that the reason behind it is that since tasks (b, c, d) include more motion in the arm, the robot trajectory provides broader observation ranges, hence, the particle filter can resolve the redundancy better and ends with less deviation. Similar phenomena occurs for the estimated postures from OpenPose. During tasks (c, d) the motion of the human causes bigger movement for the OpenPose keypoints, hence the deviations in those tasks are less than tasks (a, b). Higher whiskers for estimated postures from OpenPose also shows that in some period of the time during the task, the estimated posture has been wrong while the projected anatomical landmarks on the human skeleton were close to their corresponding keypoints from OpenPose. This is due to the redundancy in projection 3D points into 2D points in the camera frame. Figure~\ref{fig:openpose_wrong_wm} shows two example of those moments. In each example, we see that the estimated posture~(pink skeleton) matches well with the MoCap posture~(green skeleton) in the video frame, while the view from another angle shows that they are very different.

When we compare the results from the multi-sensory approach with the results from our previous work~\cite{yazdani2020leader}; in which we had median of deviation less than 5 deg and upper quantile less than 15 deg while we used the limited range of motion for torso and initializing particles around the neutral posture; we can conclude that by our multi-sensory approach is slightly more accurate, while it does not need any knowledge on the range of motion used in the task or any prior information about the posture which might reduce the unreliability of our approach.
\begin{figure}[t]
% \vspace{-3pt}
\begin{centering}
% \hspace{-1cm}
\includegraphics[width=0.9\textwidth]{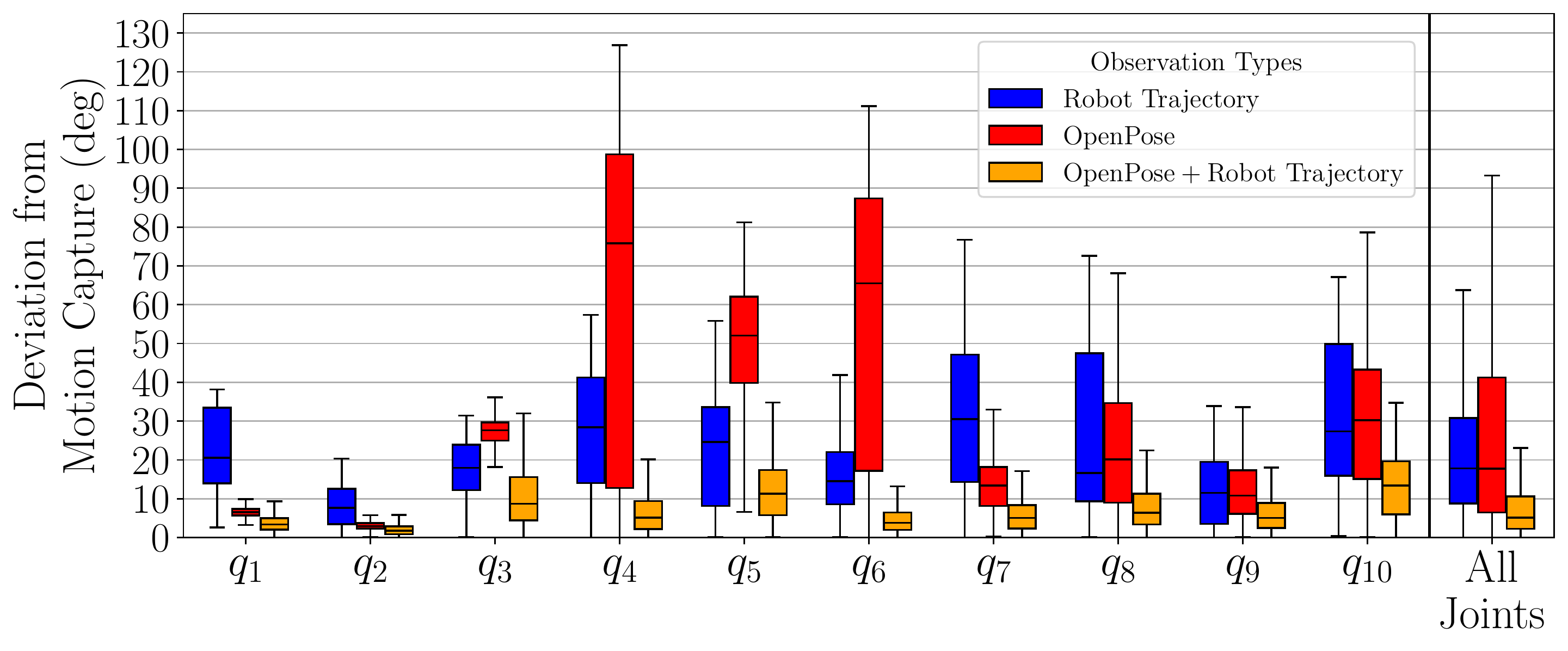}
\caption{Deviation of the posture estimated by the proposed approach from MoCap postures for all the participants and different human joints. }
\label{fig:boxplot_joints}
% \vspace{-0.5cm}
\end{centering}
\end{figure}

Next, we compare deviation of each posture estimation approach from MoCap posture among the human joints in Fig.~\ref{fig:boxplot_joints}. We see that the deviation for our multi-sensory approach is very low among all the joints, while estimated postures from OpenPose have high deviation in upper arm abduction~($q_4)$, upper arm vertical rotation~($q_5$), and upper arm horizontal rotation~($q_6$). These high deviations also match with what we discussed in Fig.~\ref{fig:openpose_wrong_wm} and the same reasoning holds here too.

\begin{figure}[t!]
\begin{centering}
\includegraphics[width=0.8\textwidth]{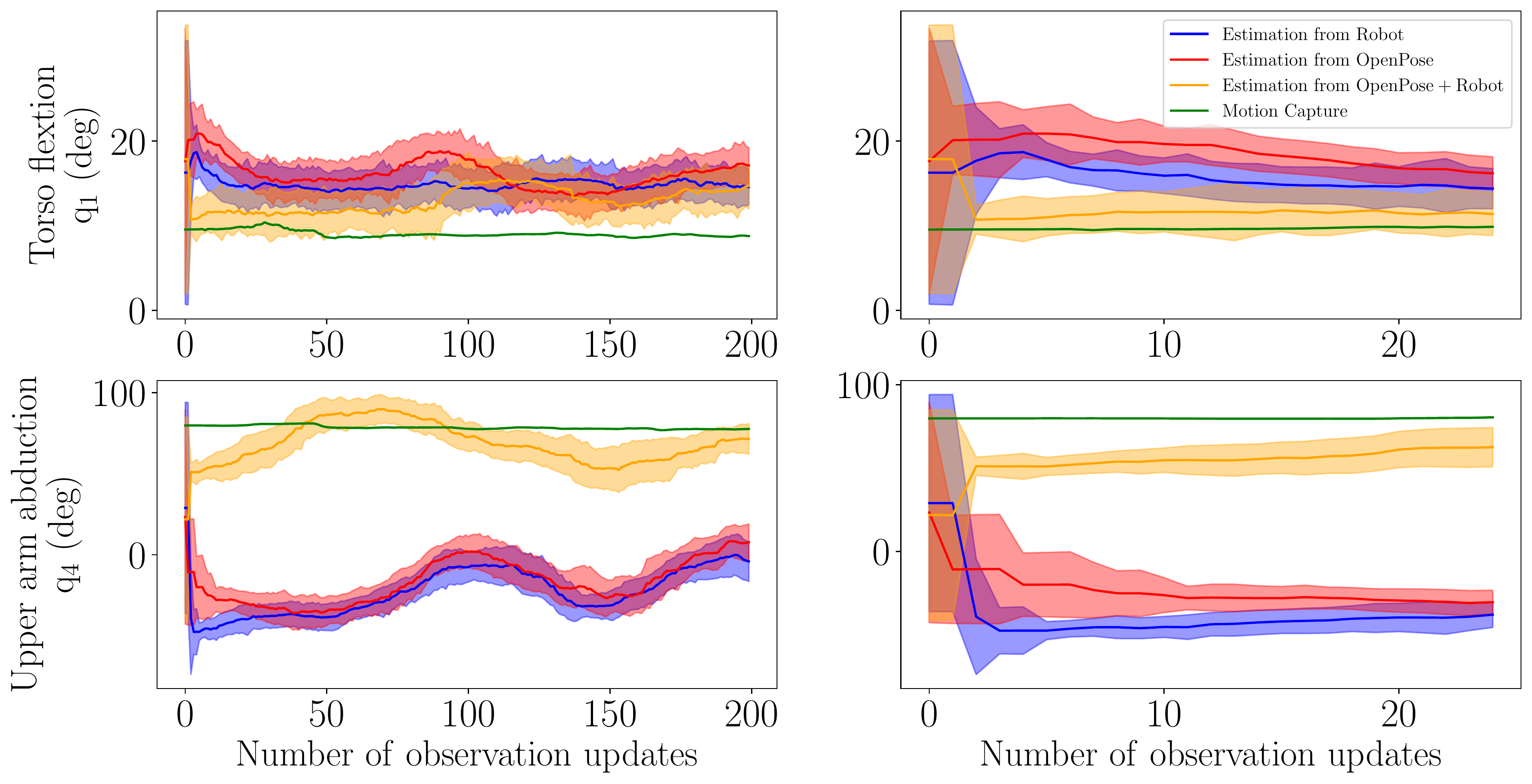}
\caption{The distribution of particles over time. The orange, red and blue lines in the figure are the average of joint angles and the shaded area around them shoes one standard deviation around the average value. The left column of the figure focuses on the full range of observation steps during the task, while the right columns shows the first 24 steps of observation.}
\label{fig:particles}
\end{centering}
\end{figure}

Finally, to show how the particles for two joints change through time, we plot the distribution of the joint angle of particles for a subject performing task (a) in Fig.~\ref{fig:particles}. Focusing on the initial 5 steps, we see that the standard deviation of the particles' joint angles from the multi-sensory approach shrinks faster than the other two approaches, which means that the particles converged to \textit{a solution} faster than the other two approaches. Also note that the converged \textit{solution} from the multi-sensory approach is closer to the estimated joint angles from the MoCap~(green line).

\begin{figure}[t]
% \vspace{-3pt}
\begin{centering}
% \hspace{-1cm}
\includegraphics[width=0.9\textwidth]{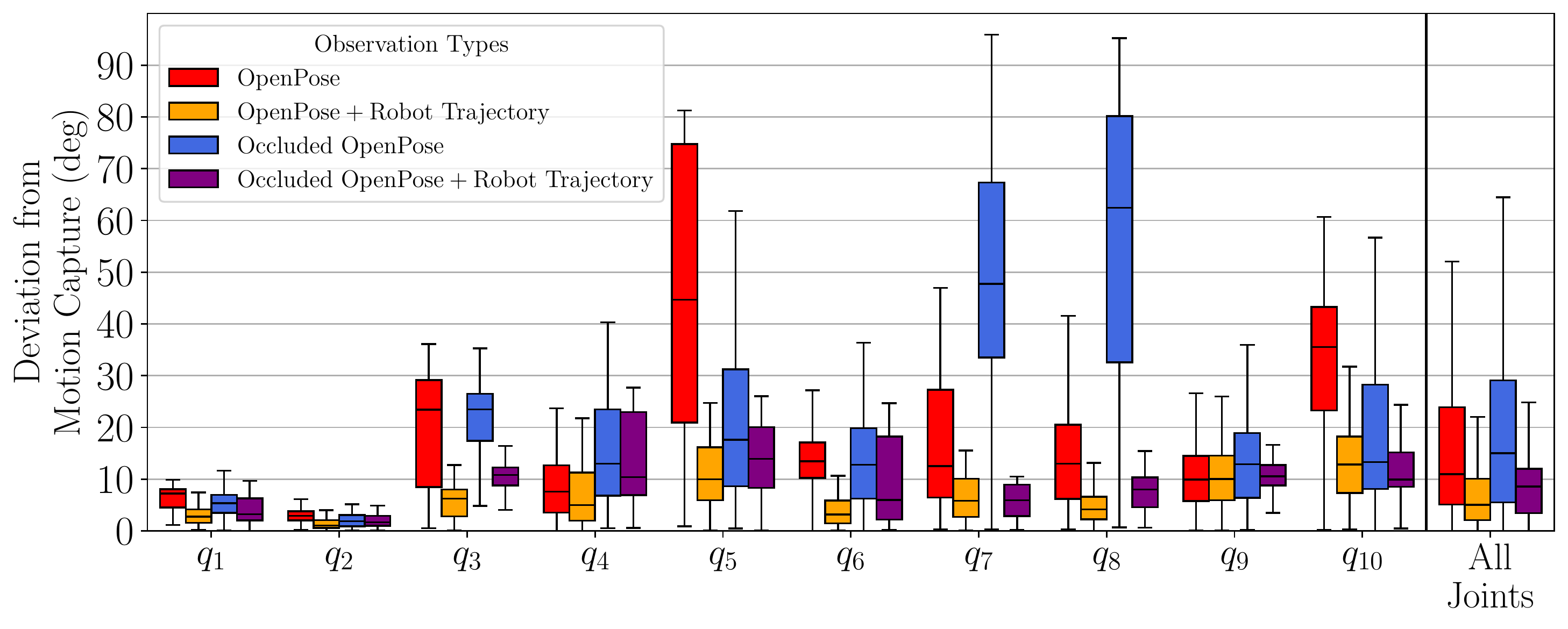}
\vspace{-4mm}
\caption{Deviation of the postures estimated by the proposed approaches from MoCap postures in occluded and non-occluded scenarios. The plot is for one subject performing task (c). }
\label{fig:occlusion_plot}
% \vspace{-0.5cm}
\end{centering}
\end{figure}

\subsection{Robustness to Occlusion}
We synthetically occluded video frames of a scenario to evaluate the robustness of the OpenPose-based posture estimation approaches. Figure~\ref{fig:occlusion_plot} visualizes the deviation of estimated postures from motion capture for each joint for the scenario with occlusion and without occlusion. The plot shows that in the multi-sensory posture estimation approach from the robot and the OpenPose has slightly higher deviation compared to the same approach in non-occluded scenario. This reveals the robustness to the occlusion for our approach compared to the approach solely using OpenPose.

\subsection{Risk Assessment}
\begin{figure}[t!]
\begin{centering}
\includegraphics[width=0.7\textwidth]{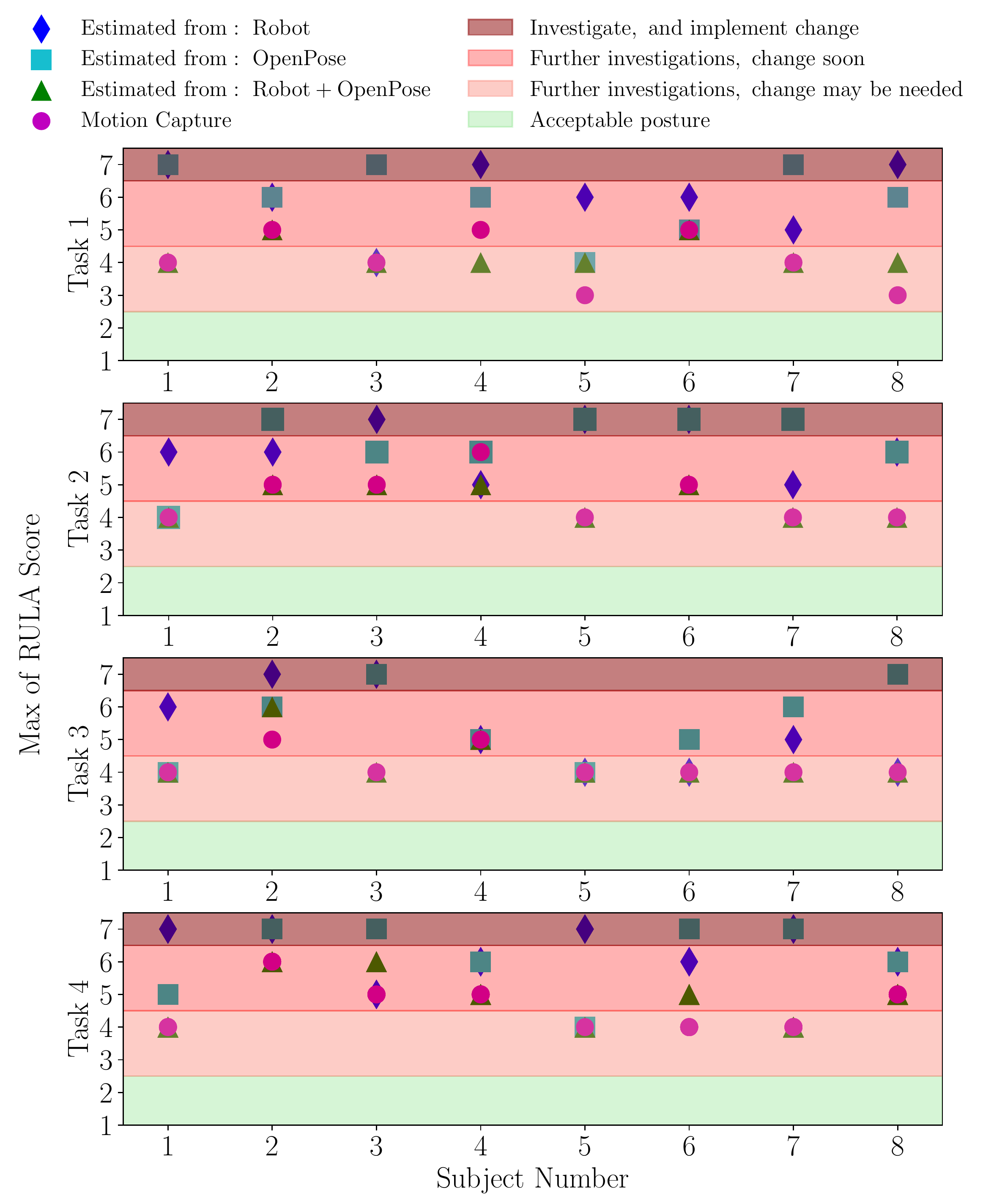}
\caption{Comparison of the maximum values of RULA scores for tasks using the estimated postures from the 3 approaches, and MoCap postures for all the participants.}
\label{fig:rula_subject}
\end{centering}
\end{figure}

Fig.~\ref{fig:rula_subject} shows the maximum RULA score during the tasks~(the one most often used in ergonomics) for the postures estimated from the three approaches, and MoCap postures. We also summarized them in Table~\ref{tab:rula_results}. All the approaches were successful in identifying all instances where the RULA score was higher than 2, i.e. place where ``further investigation or change may be or is needed''. Results prove that the multi-sensory approach outperforms the two alternative approaches and it the accuracy of it is better that the approach from our previous work in which we had the observation from the robot with limited range of motion for torso and initializing the particles around the neutral posture. Once more, it proves that using the extra camera and 2D OpenPose predictions as the additional observation removes the requirement for having prior information about the task in posture estimation.
\begin{table}[t]
  \caption{Summary of the deviation of the maximum of RULA score for all tasks and subjects. }
  \label{tab:freq}
  \begin{tabular}{lcc}
    \toprule
    Posture estimation approach&Same max (RULA score)&Same interpretation of max (RULA score)\\
    \midrule
    From Robot & 18.75\% & 31.25\%\\
    From OpenPose & 21.88\% & 40.62\%\\
    From Robot+OpenPose & 75\% & 87.5\%\\
    From Robot (with torso limitation)~\cite{yazdani2020leader} & 65.63\%& 84.37\%\\
  \bottomrule
\end{tabular}
\label{tab:rula_results}
\end{table}

%%%%%%%%%%%%%%%%%%%%%%%%%%%%%%%%%%%%%%%%%%%%%%%%%%%%%%%%%%%
\section{Conclusion and Future Directions}
In this paper, we proposed a multi-sensory 3D posture estimation approach for teleoperation, which is low-cost and robust to occlusion and achieving high quality redundancy resolution. In our approach we used the observations from 2D postures predicted by OpenPose over the video frames of a single RGB camera, and the trajectory of the leader robot while performing a teleoperation task. We modeled the problem as a partially-observable dynamical system and inferred the 3D posture using a particle filter. We compared our results with the estimated postures solely from the OpenPose and the estimated postures solely from the leader robot's trajectory, as well as their performance in ergonomic postural assessment, through a human subject study.
Our results show that the estimated postures from our multi-sensory approach have less deviation from the gold-standard MoCap postures. Moreover, the observations from two sensor highly improves the redundancy resolution and makes it robust to segment occlusions. 

In this paper, we focused on teleoperation applications, however our approach can be used for other types of physical human-robot applications with minor modification.  We would like to extend our approach to be used in virtual reality~(VR) industry to estimates the full posture of the human based on the sensory information available from VR headset and controllers and a simple camera that could be a webcam or a phone camera. In addition, we want to investigate the hypothesis that including the DULA and DEBA models for human comfort from~\cite{yazdani2022dula} in our postural estimation as a weighting function for particles can improve the redundancy resolution and accuracy of the estimated postures. Moreover, the probabilistic distribution of estimated postures and their corresponding ergonomic risk scores can be used in human-aware planning for shared autonomy and other pHRI applications.

%%
%% The acknowledgments section is defined using the "acks" environment
%% (and NOT an unnumbered section). This ensures the proper
%% identification of the section in the article metadata, and the
%% consistent spelling of the heading.
\begin{acks}
Research reported in this publication was supported in part by DARPA under grant N66001-19-2-4035 and the National Institute for Occupational Safety and Health under award number T420H008414-10. The subject study in this paper is approved by The University of Utah Institutional Review Board~(IRB\textunderscore00094626).
\end{acks}

%%
%% The next two lines define the bibliography style to be used, and
%% the bibliography file.
\bibliographystyle{ACM-Reference-Format}
\bibliography{draft}

%%
%% If your work has an appendix, this is the place to put it.
% \appendix

% \section{Research Methods}

% \subsection{Part One}

% Lorem ipsum dolor sit amet, consectetur adipiscing elit. Morbi
% malesuada, quam in pulvinar varius, metus nunc fermentum urna, id
% sollicitudin purus odio sit amet enim. Aliquam ullamcorper eu ipsum
% vel mollis. Curabitur quis dictum nisl. Phasellus vel semper risus, et
% lacinia dolor. Integer ultricies commodo sem nec semper.

% \subsection{Part Two}

% Etiam commodo feugiat nisl pulvinar pellentesque. Etiam auctor sodales
% ligula, non varius nibh pulvinar semper. Suspendisse nec lectus non
% ipsum convallis congue hendrerit vitae sapien. Donec at laoreet
% eros. Vivamus non purus placerat, scelerisque diam eu, cursus
% ante. Etiam aliquam tortor auctor efficitur mattis.

% \section{Online Resources}

% Nam id fermentum dui. Suspendisse sagittis tortor a nulla mollis, in
% pulvinar ex pretium. Sed interdum orci quis metus euismod, et sagittis
% enim maximus. Vestibulum gravida massa ut felis suscipit
% congue. Quisque mattis elit a risus ultrices commodo venenatis eget
% dui. Etiam sagittis eleifend elementum.

% Nam interdum magna at lectus dignissim, ac dignissim lorem
% rhoncus. Maecenas eu arcu ac neque placerat aliquam. Nunc pulvinar
% massa et mattis lacinia.

\end{document}